\documentclass[letterpaper, 10 pt, conference]{ieeeconf}  

\IEEEoverridecommandlockouts                              

\usepackage{url}
\usepackage{xcolor}
\usepackage{cite}

\usepackage{subfig}
\usepackage{tikz}
\usepackage{pgfplots}
\usepackage{tabularx}
\usepackage{amsmath}
\usepackage{graphicx}

\usepackage{hyperref}
\usepackage{amssymb}
\usepackage{amsmath}
\usepackage[english]{babel}
\usepackage[nolist,nohyperlinks]{acronym}
\usepackage{todonotes}
\usepackage{epstopdf}
\usepackage{bm}  
\usepackage[linesnumbered]{algorithm2e} 

\hypersetup{
    colorlinks=true,
    linkcolor=blue,    
    urlcolor=blue,
}
\makeatletter
\newcommand{\removelatexerror}{\let\@latex@error\@gobble}
\makeatother

\newlength\algowd

\let\oldnl\nl
\newcommand{\nonl}{\renewcommand{\nl}{\let\nl\oldnl}}

\newcommand{\codecomment}[1]{\,\,{\ttfamily\scriptsize // #1}}

\title{\LARGE \bf
3D Multi-Robot Exploration with a Two-Level Coordination \\ Strategy and Prioritization
}

\author{Luigi Freda$^{1}$ \and
Tiago Novo$^{2}$ \and 
David Portugal$^{2}$ \and 
Rui P. Rocha$^{2}$
\thanks{$^{1}$
        {\tt\small \{luigifreda\}@gmail.com}}%
\thanks{$^{2}$Institute of Systems and Robotics of the University of Coimbra, 3030-290 Coimbra, Portugal
        {\tt\small \{rprocha, davidbsp\}@isr.uc.pt}}%
        }

\begin{document}

\maketitle
\thispagestyle{empty}
\pagestyle{empty}

\begin{abstract}
This work presents a 3D multi-robot exploration framework for a team of UGVs moving on uneven terrains. The framework was designed by casting the two-level coordination strategy presented in \cite{fre2018patrolling} into the context of multi-robot exploration. The resulting distributed exploration technique minimizes and explicitly manages the occurrence of conflicts and interferences in the robot team. 
Each robot selects where to scan next by using a receding horizon next-best-view approach~\cite{bircher2016receding}. A sampling-based tree is directly expanded on segmented traversable regions of the terrain 3D map to generate the candidate next viewpoints. 
During the exploration, users can assign locations with higher priorities on-demand to steer the robot exploration toward areas of interest.
The proposed framework can be also used to perform coverage tasks in the case a map of the environment is a priori provided as input. An open-source implementation is available online.

\end{abstract}

\section{Introduction}
\begin{figure}[!h]
\begin{center}   
     \subfloat[\label{teaser1}]{%
       \includegraphics[height=0.44\textwidth]{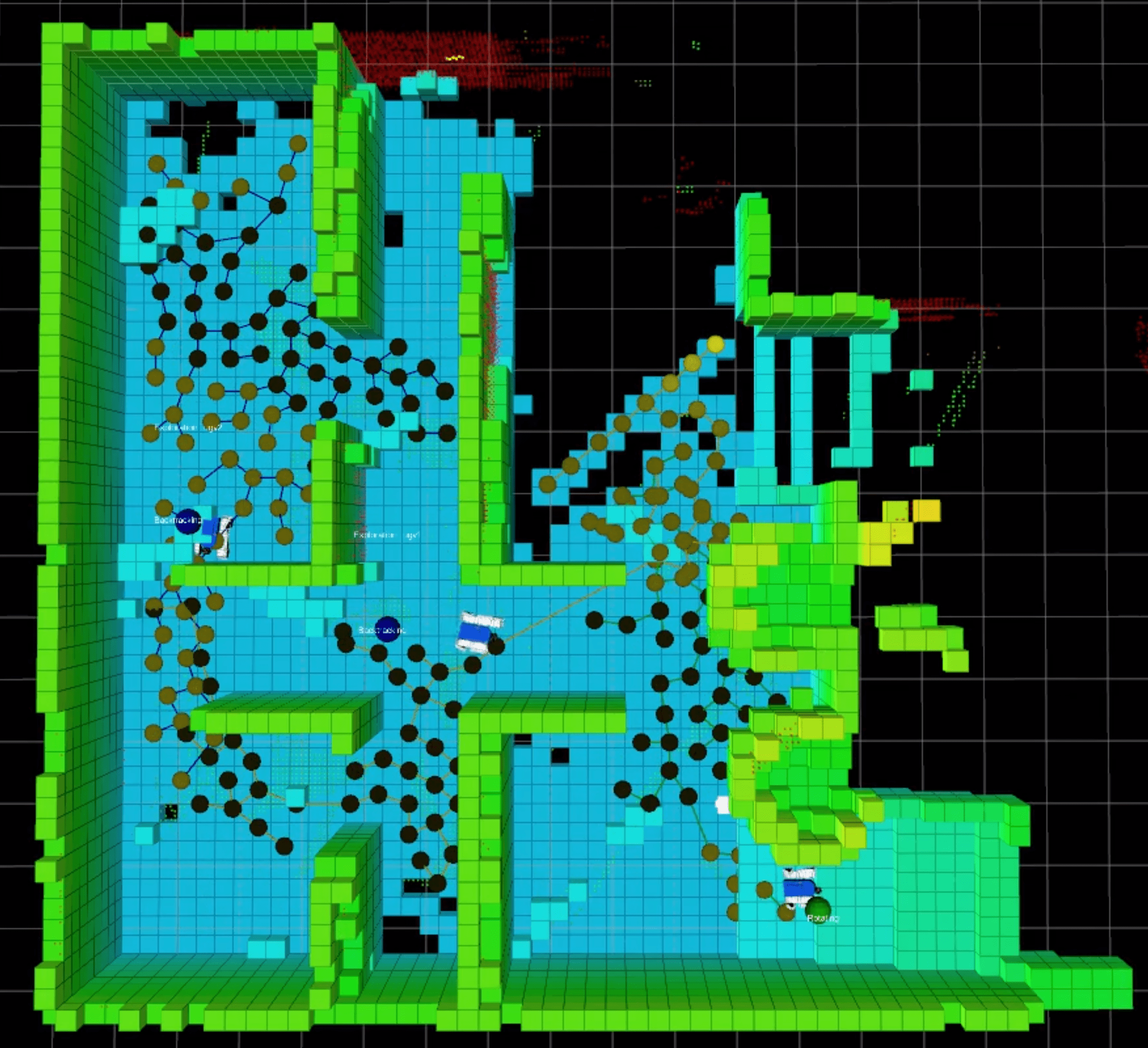}
     } \\         
     \subfloat[\label{teaser2}]{%
       \includegraphics[height=0.173\textwidth]{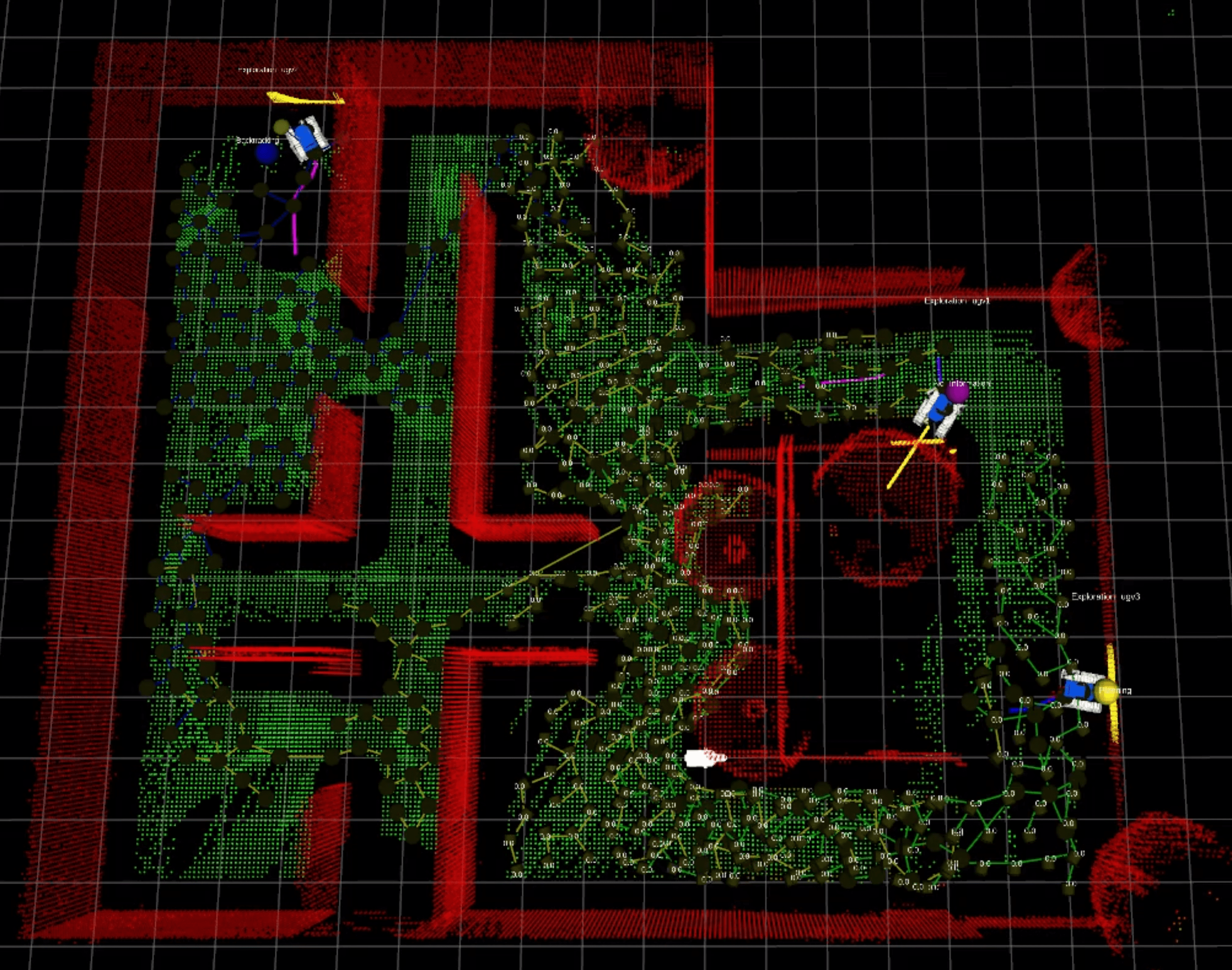}
     }  
     \quad
     \subfloat[\label{teaser3}]{%
       \includegraphics[height=0.173\textwidth]{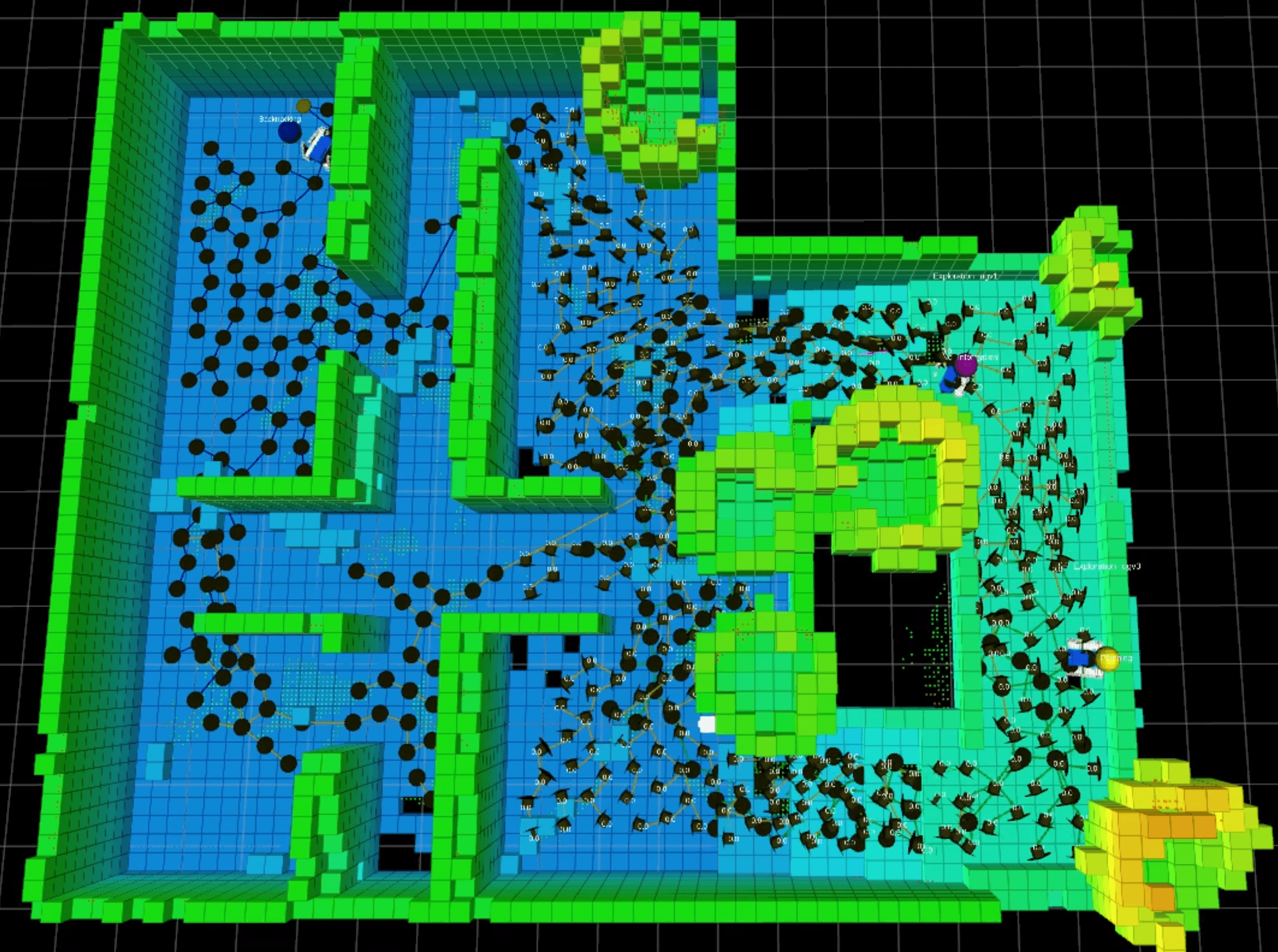}
     }           
     \caption{\textit{(a)} The volumetric map and the frontier tree built by a team of 3 robots. The volumetric map represents free and occupied space and is used to compute the information gain of each candidate next viewpoint. Nodes with high information gain are inserted into the frontier tree and updated for planning. \textit{(b)} and \textit{(c)}: The traversability and volumetric maps obtained at the end of the exploration process. The traversability map is used to expand a search tree of candidate viewpoints and generate new safe paths at any given time.}    
     \label{Fig:TeaserFigure} 
\end{center}
\end{figure}

In the near future, fleets of autonomous robots will be able to flexibly and cooperatively perform multiple tasks such as exploration, coverage and patrolling~\cite{Kaminka2008,portugal2017performance}. Amongst these tasks, an exploration mission is crucial for first assessments and to preliminarily build a model of an unknown environment. This operation typically requires a higher level of autonomy and robustness, especially with UGVs or UAVs operating in complex 3D 
environments~\cite{kruijff2016deployment,surmann2003autonomous,bircher2016receding,papachristos2017uncertainty, shen2012stochastic}. 

Specifically, the goal of a team of exploring robots is to cooperatively cover an unknown environment with sensory perceptions~\cite{FrOrVe:09}. Typically, the expected output of exploration is a 3D map of the environment~\cite{surmann2003autonomous} or the discovery of interesting information/objects~\cite{DaDaGkLePaBa2017,aydemir2013active}.
Indeed, multi-robot exploration has a wide range of potential applications, spanning from search-and-rescue operations~\cite{colas20133d,murphy2014disaster,kleiner2016special} to surveillance~\cite{bhattacharya2007motion}, mapping~\cite{dube2017online} and planetary missions.

In general, a multi-robot system presents many advantages over a single robot system~\cite{yan2013survey}: completion time is reduced, information sharing and redundancy can be used to increase the map accuracy and localization quality~\cite{rekleitis2001multi, dube2017online}. Nonetheless, taking advantage of a multi-robot architecture and actually attaining performance improvements requires the design of sophisticated strategies which can guarantee \emph{cooperation} (avoid useless actions) and \emph{coordination} (avoid conflicts). 
In the realm of UGVs, these strategies are particularly crucial. Indeed, narrow passages (due for example to collapsed infrastructures or debris~\cite{kruijff2016deployment,murphy2014disaster}) typically generate spatial conflicts amongst teammates. A suitable strategy is required to \emph{(i)} minimize interferences and \emph{(ii)} recognize and resolve possible incoming deadlocks. Neglecting the occurrence of such events can hinder UGVs activities or even provoke major failures.  

A team of exploring UGVs has to face many challenges. For instance, the UGVs are required to navigate over a 3D uneven and complex terrain. Moreover, the environment may be dynamic and large-scale~\cite{Cadena16tro-SLAMfuture}. In this case, UGVs must continuously and suitably update their internal representations of the surrounding scenario to robustly localize and to best adapt their behaviors to environment changes. 
Last but not least, in order to collaborate, UGVs must continuously exchange coordination messages and share their knowledge over an unreliable network infrastructure. The design of a robust communication protocol is crucial.  

In this paper, we address the aforementioned challenges. We present a multi-robot exploration framework designed by casting the two-level coordination strategy presented in \cite{fre2018patrolling} into a 3D exploration context. The resulting distributed technique minimizes and explicitly manages the occurrence of conflicts and interferences in the team. Each robot selects where to scan next by using a receding horizon next-best-view approach~\cite{bircher2016receding}. Here, a sampling-based tree is directly expanded on segmented traversable regions of the terrain 3D map (see Fig.~\ref{Fig:TeaserFigure}).
During the exploration, users can assign locations with higher priorities on-demand to steer the exploration toward areas of interest. The proposed framework can be also used to perform coverage tasks in the case a 3D map of the environment is provided a priori as input.
We evaluate our system through both simulation analyses and real-world experiments. An open-source implementation is available at \href{https://github.com/luigifreda/3dmr}{github.com/luigifreda/3dmr}.

The remainder of this paper is organized as follows. First, Section~\ref{Sect:RelatedWorks} presents and discusses related works. Then, Section~\ref{Sect:ProblemSetup} describes the exploration problem. Next, an overview of our approach is presented in Section~\ref{Sect:Method}. Our two-level exploration strategy is then described in Sections~\ref{Sect:TwoLevelCoordinationStrategy} and~\ref{Sect:ExplorationAgent}. Subsequently, Section~\ref{Sect:Coverage} explains how our framework can be used for coverage tasks. Results are presented in Section~\ref{Sect:Results}. Finally, Section~\ref{Sect:Conclusions} presents our conclusions.

\section{Related Works}\label{Sect:RelatedWorks}

In the literature, the problem of covering a scene with sensory perceptions comes in many flavors. Essentially, a sequence of viewpoints must be planned in order to gather sensory perceptions over a scene of interest. Three main specializations can be considered for this \emph{viewpoint planning} problem: next-best-view, exploration and coverage. 

\subsection{Next-best-view}

If the scene consists of a single-object without obstacles, \emph{next-best-view} planning algorithms are in order~\cite{connolly1985determination, vasquez2014volumetric,dunn2009developing, krainin2011autonomous}. In this case, the goal is to obtain an accurate 3D reconstruction of an arbitrary object of interest. These approaches typically sample candidate view configurations within a sphere around the target object and select the view configuration with the highest information gain. The downside of these strategies is they do not scale well in multi-objects scenes. 

In this work, the proposed exploration framework uses the same philosophy of next-best-view approaches. First, candidate views are generated. In our case, we locally expand a sampling-based tree of candidate views over the traversable terrain surrounding the robot. Then, the next best view is selected. As in~\cite{bircher2016receding}, we first identify the branch $b^*$ of the expanded tree which maximizes the total information gain (as in a receding-horizon scheme) and then select as next view the nearby node along $b^*$ (as in a motion-predictive approach). 

\subsection{Exploration}

When the scene is an unknown environment, viewpoint planning is referred to as \emph{exploration}. 

The vast majority of exploration strategies are \emph{frontier-based}~\cite{Ya:98}. Here, the \emph{frontier} is  defined as the boundary between known and unknown territory and is approached in order to maximize the expected information gain. In most strategies, a team of robots cooperatively build a metric representation of the environment, where frontier segments are extracted and used as prospective viewpoints.

In~\cite{BuMoStSc:05}, Burgard et al. presented a frontier-based decentralized strategy where robots negotiate targets by optimizing a mixed utility function which takes into account the expected information gain, the cost-to-go and the number of negotiating robots.
In~\cite{HoPaSu:06}, the same decentralized frontier-based approach extended to a large-scale heterogeneous team of mobile robots.

An incremental deployment algorithm is presented in~\cite{HoMaSu:03}, where robots approach the frontier while retaining visual contact with each other.

In~\cite{freda2005frontier,Franchi-2009}, a sensor-based random graph is expanded by the robots by using a randomized local planner that automatically performs a trade-off between information gain and navigation cost.

A interesting class of exploration methods fall under the hat of \emph{active SLAM}~\cite{leung2006active,kim2015active,carlone2014active,valencia2018active} (or integrated exploration~\cite{MaWiBoDu:02}), which considers the correlation between viewpoint selection and the onboard localization/mapping quality. For instance, in~\cite{KoStFoKoLi:03, freda2006randomized}, the utility of a particular frontier region is also considered from the viewpoint of relative robot localization. Similarly, \emph{belief-space planning} was proposed to integrate the expected robot belief into its motion planning~\cite{costante2016perception,indelman2015planning}.

An interesting multi-robot architecture is presented in~\cite{ZlStDiTh:02}, where robots are guided through the exploration by a market economy. Similarly, in~\cite{SiApBuFoMoThYo:00}, a centralized frontier-based approach is proposed in which a bidding protocol is used to assign frontier targets to the robots.

In~\cite{shade2011choosing}, an exploration strategy is driven by the resolution of a partial differential equation. A similar concept is presented in~\cite{shen2012stochastic}. Here, in order to solve a stochastic differential equation, Shen et al. use particles as an efficient sparse representation of the free space and for identifying frontiers.

Biological-inspired strategies based on Particle Swarm Optimization (PSO) are presented in~\cite{couceiro2011novel}, where an exploration task is defined through the distributed optimization of a suitable sensing function. 

In~\cite{osswald2016speeding}, an efficient exploration strategy exploits background information (i.e. a topo-metric map) of the environment in order to improve time performances.

\subsection{Coverage}

Finally, when a prior 3D model of the scene is assigned, the viewpoints planning problem is known as \emph{coverage}~\cite{choset2001coverage,hazon2008redundancy}. In~\cite{Kaminka2008}, a special issue collected new research frontiers and advancements on multi-robot coverage along with multi-robot search and exploration. In~\cite{agmon2008giving}, Agmon et al. proposed a strategy for efficiently building a coverage spanning tree for both online and offline coverage, aiming to minimize the time to complete coverage.  In~\cite{heng2015efficient}, an efficient frontier-based approach was proposed to solve at the same time the problems of exploration (maximize the size of the reconstructed model) and coverage (observe the entire surface of the environment, maximize the model completeness). In~\cite{galceran2015coverage,englot2013three}, coverage strategies were presented for inspecting complex structures on the ocean floor by using autonomous underwater vehicles.


\section{Problem Setup}\label{Sect:ProblemSetup}
\begin{table}
\begin{center} 
\caption{Table of symbols.}
\label{Tab:Symbols}
\begin{tabular}{llll}
Symbol & Description  \\
\cline{1-2}\\[-1.0em]
${\cal W}$ & Environment\\
$T$ & Time interval $[t_0,\infty)$ \\
${\cal S}$ & Surface terrain in ${\cal W}$\\
${\cal O}$ & Obstacle region\\
${\cal C}$ & Configuration space of each robot \\
${\cal A}_j(\bm{q})$ & Region occupied by robot $j$ at configuration $\bm{q} \in {\cal C}$ \\
${\cal V}_j(\bm{q})$ & View of robot $j$ at $\bm{q}$ \\
${\cal E}_j^k$ & Explored region of robot $j$ at step $k$ \\
${\cal E}^k$ & Region explored by the robot team at step $k$ \\
${\cal R}^k_j$ & Exploration-safe region of robot $j$ at step $k$ \\
${\cal Q}_j^k$ & Informative-safe region of robot $j$ at step $k$ \\
\cline{1-2}
\end{tabular}
\end{center}
\end{table}

A team of $m \geq 2$ ground exploration robots are deployed in an unknown environment and are constrained to move on an uneven terrain. The team has to perform an exploration mission, i.e. cooperatively cover the largest possible part of the environment with sensory perceptions~\cite{FrOrVe:09}. The output of the exploration process is a 3D model of the environment.

\subsection{Notation}

In the remainder of this paper, we will use the following convention: Given a symbol $a^k_j$, unless otherwise specified, the superscript $k$ denotes a time index while the subscript $j$ refers to robot $j$. For instance, $\bm{q}_j^k$ is the configuration of robot $j$ at exploration step $k$. 

A list of the main symbols introduced hereafter is reported in Table~\ref{Tab:Symbols}.

\subsection{3D Environment, Terrain and Robot Configuration Space}\label{Sect:3DEnvModel}

Let $T=[t_0,\infty) \subset \mathbb{R}$ denote a \emph{time interval}, where $t_0 \in \mathbb{R}$ is the starting time.
The 3D \emph{environment} ${\cal W}$ is a compact connected region in $\mathbb{R}^3$. 
The obstacle region is denoted by ${\cal O} \subset \mathbb{R}^3$. For ease of presentation, we assume ${\cal O}$ to be static for the moment\footnote{In our experiments, we relaxed this assumption by considering an environment populated by low-dynamic objects~\cite{walcott2012dynamic}. This is possible by representing the environment with a suitable dynamic model.}.

The robots move on a 3D terrain, which is identified as a compact and connected manifold ${\cal S}$ in ${\cal W}$. The \emph{free environment} is ${\cal W}_{\rm
free}={\cal W} \setminus \big( {\cal O} \cup {\cal S} \big)$.

The configuration space $\cal C$ of each robot is the special Euclidean group $SE(3)$~\cite{Lavalle-2006}. In particular, a robot configuration $\bm{q} = [\bm{p},\bm{\phi}]^T \in {\cal C}$ consists of a 3D position $\bm{p}$ of the robot representative centre and a 3D orientation $\bm{\phi}$. ${\cal A}_j(\bm{q}) \subset \mathbb{R}^3$ denotes the compact region occupied by robot~$j$ at $\bm{q} \in {\cal C}$.

A robot configuration $\bm{q} \in {\cal C}$ is considered \emph{valid} if the robot at $\bm{q}$ is safely placed over the 3D terrain $\cal S$. This requires $\bm{q}$ to satisfy some validity constraints defined according to~\cite{hait2002algorithms}.
 
A robot path is a continuous function $\bm{\tau}: [0,1] \to {\cal C}$.
A path $\bm{\tau}$ is \emph{safe} for robot $j$ if for each $s \in [0,1]$: ${\cal A}_j(\bm{\tau}(s)) \cap {\cal O} = \emptyset$ and $\bm{\tau}(s) \in {\cal C}$ is a valid configuration. 

We assume each robot in the team is \emph{path controllable}, that is each robot can follow any assigned safe path in $\cal C$ with arbitrary accuracy~\cite{Franchi-2009}.

\begin{figure}[!t]
\begin{center}
	\includegraphics[width=\linewidth]{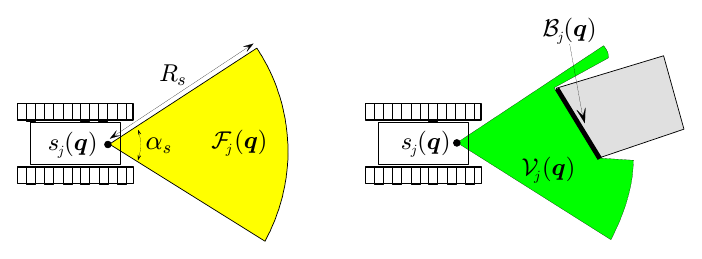} \caption{{\em Left}: robot $j$ with its sensor center $s_j(\bm{q})$ and the associated field of view ${\cal F}_j(\bm{q})$ at configuration $\bm{q}$. {\em
Right}: the view ${\cal V}_j(\bm{q})$ and the visible obstacle boundary
${\cal B}_j(\bm{q})$.} \label{Fig:FOVandPerception}
\end{center}
\end{figure}

\subsection{Sensor Model}

Each robot of the team is equipped with a 3D laser range-finder rigidly attached to its body. Teammates are able to localize in a common global map frame (see~\cite{fre2018patrolling}). We formalize the laser sensor operation as follows.

Assuming that robot $j$ is at
$\bm{q}$, denote by ${\cal F}_j(\bm{q}) \subset \mathbb{R} ^N$ the compact region
occupied by its sensor \emph{field of view}, which is
\emph{star-shaped} with respect to its  \emph{sensor center}
$s_j(\bm{q})\in{\cal W}$.  In $\mathbb{R} ^2$, for instance, ${\cal F}_j(\bm{q})$
can be a circular sector with apex $s_j(\bm{q})$, opening angle
$\alpha_s$ and radius $R_s$, where the latter is the perception range (see Fig.~\ref{Fig:FOVandPerception}, left). 

With robot $j$ at $\bm{q}$, a point $\bm{p} \in {\cal W}$ is said to be
\emph{visible from the sensor} if $\bm{p} \in {\cal F}_j(\bm{q})$ and
the open line segment joining $\bm{p}$ and $s_j(\bm{q})$ does not intersect
\mbox{$\partial {\cal O} \cup {\cal S} \cup \partial {\cal A}_j(\bm{q})$}. 
At each configuration $\bm{q}$, the robot sensory
system returns a 3D scan with the following information (see Fig.~\ref{Fig:FOVandPerception}, right):

\begin{itemize}

\item the \emph{visible free region} (or \emph{view}) ${\cal V}_j(\bm{q})$, i.e. all the points of ${\cal W}_{\rm free}$ that are visible from the sensor of robot $j$;

\item the \emph{visible obstacle boundary} ${\cal
B}_j(\bm{q})= \big( \partial {\cal O} \cup {\cal S} \big) \cap \partial {\cal V}_j(\bm{q})$, i.e., all
points of $\partial {\cal O} \cup {\cal S}$ that are visible from the sensor.

\end{itemize}

The above sensor is an idealization of a `continuous' range finder. In our case, it is a 3D laser
range finder, which returns the distance to the nearest obstacle
point along the directions (\emph{rays}) contained in its field of
view (with a certain resolution). Indeed, other sensory systems, such a stereoscopic camera,
satisfy the above description.

\subsection{Exploration task}\label{Sect:ExplorationTasks}

Robot $j$ explores the world through a sequence of \emph{view-plan-move}
actions. In order to simplify the notation, we assume the robots synchronize their actions, i.e. a step $k$ identifies a common time frame $t_k \in T$ for all the robots. The following presentation can be easily adapted to the general case.

Each configuration where a view is acquired is called a
\emph{view configuration}. Let $\bm{q}_j^0$ be the initial configuration of robot $j$. Denote by $\bm{q}_j^1,\bm{q}_j^2,...,\bm{q}_j^k$
the sequence of view configurations planned by robot $j$ up to the $k$-th exploration
step. When the exploration starts, all the initial robot $j$ endogenous
knowledge  can be expressed as:
\begin{equation}\label{Eq:BootstrapExploredRegion}
{\cal E}_j^0 = {\cal A}_j(\bm{q}_j^0) \cup {\cal V}_j(\bm{q}_j^0),
\end{equation}
where ${\cal A}_j(\bm{q}_j^0)$ represents the free volume\footnote{Often,
${\cal A}_j(\bm{q}_j^0)$ in~(\ref{Eq:BootstrapExploredRegion}) is replaced by
a larger free volume $\widetilde{{\cal A}}_j$ whose knowledge comes
from an external source. This may be essential for planning safe
motions in the early stages of an exploration.} that robot $j$ body
occupies (computed on the basis of proprioceptive sensors) and
${\cal V}_j(\bm{q}_j^0)$ is the view at $q_j^0$ (provided by the exteroceptive
sensors). At step $k \geq 1$, the \emph{explored region} of robot $j$ is:
\[
{\cal E}_j^k = {\cal E}_j^{k-1} \cup {\cal V}_j(\bm{q}_j^k).
\]
At each step $k$, ${\cal E}_j^k \subseteq {\cal W}_{\rm free}$ is the
current estimate of the free world that robot $j$ has built. Since safe planning requires
${\cal A}_j(\bm{q}^k) \subset{\cal E}_j^{k-1}$ for any $k$, we have:
\begin{equation}\label{Eq:ExploredRegionDecomposition}
{\cal E}_j^k = {\cal A}_j(\bm{q}_j^0) \cup \left( \bigcup\limits_{i = 0}^k
{{\cal V}_j(\bm{q}_j^i) } \right).
\end{equation}
If we consider the whole robot team, the overall explored region at step $k$ is:
\begin{equation}\label{Eq:TotalExploredRegionDecomposition}
{\cal E}^k =  \bigcup\limits_{j = 1}^m
{{\cal E}_j^k } = \bigcup\limits_{j = 1}^m {\cal A}_j(\bm{q}_j^0) \cup \left( \bigcup\limits_{j = 1}^m \bigcup\limits_{i = 0}^k
{{\cal V}_j(\bm{q}_j^i) } \right).
\end{equation}
In the above equations, the union operator $\cup$ denotes a data-integration operation which depends on the adopted spatial model (see Sect.~\ref{Sect:VolumetricMapping}).  

A point $\bm{p} \in {\cal W}_{\rm free}$ is defined \emph{explored} at
step $k$ if it is contained in ${\cal E}^k$ and \emph{unexplored}
otherwise. A configuration $\bm{q}$ is \emph{exploration-safe} at step $k$ if ${\cal
A}_j(\bm{q}) \subset {\rm cl}({\cal E}^k)$, where cl$(\cdot)$ indicates the
set closure operation (configurations that bring the robot in
contact with obstacles are allowed). A path $\bm{\tau}$ in $\cal C$ is \emph{exploration-safe} for robot $j$ at step $k$ if for each $s \in [0,1]$ the configuration $\bm{\tau}(s) \in {\cal C}$ is exploration-safe.

The \emph{exploration-safe region} 
${\cal R}_j^k$
of robot $j$ collects all the configurations that
are reachable from $\bm{q}_j^0$ through an exploration-safe path at step $k$. Indeed, ${\cal
R}_j^k$ represents a configuration space image of
${\cal E}^k$ for robot~$j$. 
 
The \emph{view plan} $\pi_j$ of robot $j$ is a finite sequence of view configuration $\pi_j = \{\bm{q}_j^k\}_{k=0}^{l_j}$, where $l_j$ is the length of $\pi_j$. An \emph{exploration plan} is a view plan such that $\bm{q}_j^k \in {\cal
R}_j^k$ at each step $k$.
A \emph{team exploration strategy} $\Pi = \{\pi_1,...,\pi_m\}$ collects the exploration plans of the robots in the team. Here, $l = {\rm max}(l_1,...,l_m)$ is the length of the strategy $\Pi$. 

\noindent {\bf Exploration objective}. The robots must cooperatively plan an exploration strategy $\Pi$ of minimum length $l$ such that ${\cal E}^l$ is maximized. In particular, ${\cal E}^l$ is \emph{maximized} at step $l$ if there is no robot $j$ that can plan a new view configuration $\bm{q} \in {\cal
R}_j^l$ such that ${\cal E}^l \subset ({\cal E}^l \cup {\cal V}_j(\bm{q}))$.

Other factors, such as the resulting map ``accuracy" could be taken into account in the exploration objective. 
We further develop this in the next subsection. 

\begin{figure}[!t]
\begin{center}
\begin{small}
\begin{tabular}{p{0.95\columnwidth}}
\hline\hline
   \textbf{General Exploration Method}\\
  \textbf{if} ${\cal Q}_j^k \cap {\cal D}(\bm{q}_j^k,k)  \neq \emptyset$\hfill~~~~\%{\em forwarding}\%\\
  ~~choose new $\bm{q}_j^{k+1}$  in ${\cal Q}_j^k \cap {\cal D}(\bm{q}_j^k,k)$\\
  ~~move to $\bm{q}_j^{k+1}$ and acquire sensor view\\
  \textbf{else}\hfill~~~~\%{\em backtracking}\%\\
  ~~choose visited $\bm{q}_j^b$ ($b<k$) such that ${\cal Q}_j^k \cap {\cal D}(\bm{q}_j^b,k) \neq \emptyset$ \\
  ~~move to $\bm{q}_j^b$\\
\hline\hline
\end{tabular}
\end{small}
\end{center}
\caption{The $k$-th step of a general exploration method.}
\label{Fig:GeneralExplorationStrategy}
\end{figure}

\subsection{Exploration Methods}
\label{Sect:ExplorationStrategies}

Assume that each robot can associate an information gain $I(\bm{q},k)$ to
any (safe) $\bm{q}$ at step $k$. This is an estimate of the world
information which can be discovered at the current step by acquiring
a view from $\bm{q}$.

Consider the $k$-th exploration  step, which starts with the robot $j$ at $\bm{q}_j^k$. Let ${\cal Q}_j^k \subset {\cal R}_j^k$ be the
\mbox{\emph{informative safe region}} of robot $j$, i.e. the set of
configurations which {\em (i)} have non-zero information gain, and
{\em (ii)} can be reached\footnote{The reachability requirement
accounts for possible kinematic constraints to which robot may be
subject to.} from $\bm{q}_j^k$ through a path that is exploration-safe at step $k$. A
general exploration method
(Fig.~\ref{Fig:GeneralExplorationStrategy}) searches for the next view configuration in ${\cal Q}_j^k \cap {\cal D}(\bm{q}_j^k,k)$, where
 ${\cal D}(\bm{q}_j^k,k) \subseteq {\cal C}$ is
a compact \emph{admissible set} around $\bm{q}_j^k$ at step $k$, whose size
determines the scope of the search. For example, if ${\cal D}(\bm{q}_j^k,k)
= {\cal C}$, a global search is performed, whereas the search is
local if ${\cal D}(\bm{q}_j^k,k)$ is a neighborhood~of~$\bm{q}_j^k$.

If ${\cal Q}_j^k \cap {\cal D}(\bm{q}_j^k,k)$ is not empty, $\bm{q}_j^{k+1}$ is
selected in ${\cal Q}_j^k \cap {\cal D}(\bm{q}_j^k,k)$ according to some criterion (e.g., information gain
maximization).  The robot then moves to $\bm{q}_j^{k+1}$ to acquire a new
view (\emph{forwarding}). Otherwise, the robot returns to a
previously visited $\bm{q}_j^b$ ($b<k$) such that ${\cal Q}_j^k \cap {\cal
D}(\bm{q}_j^b,k)$ is not empty (\emph{backtracking}). 

The exploration can be considered \emph{completed} at step $k$ if
${\cal Q}_j^k=\emptyset$, i.e., no informative configuration can be
safely reached. 

To specify an exploration method, one must define:
\begin{itemize}
\itemsep0em 
\item an information gain;
\item a forwarding selection strategy;
\item an admissible set ${\cal D}(\bm{q}_j^k,k)$;
\item a backtracking selection strategy.
\end{itemize}

\section{System Overview}\label{Sect:Method}
\begin{figure}[t!]
\centering
	\includegraphics[width=\linewidth]{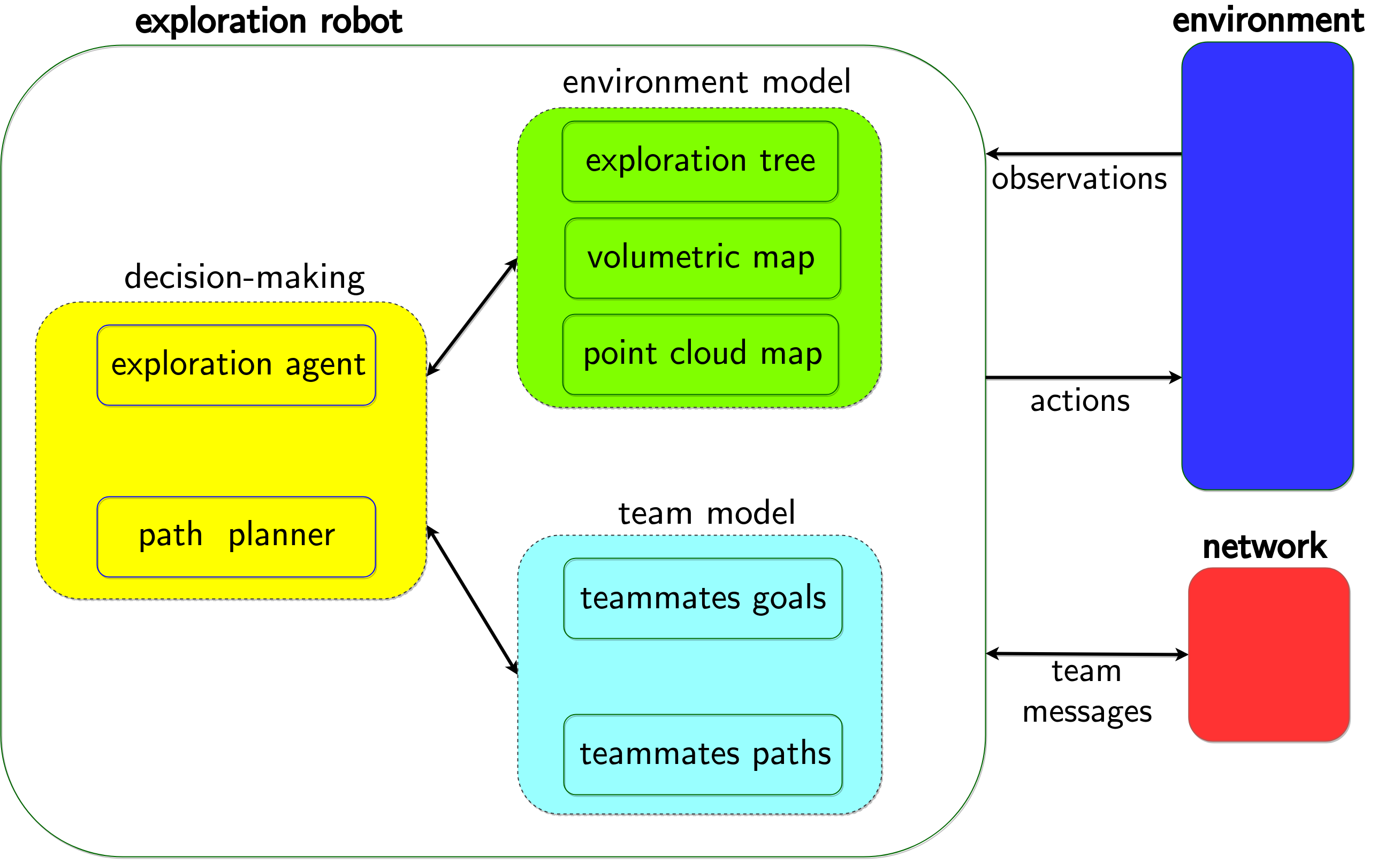}
    \caption{The exploration robot model.}%
    \label{Fig:ExplorationRobotModel}
\end{figure}

\begin{table}[!t]
\begin{center} 
\caption{Table of environment models.}
\label{Tab:EnvModels}
\begin{tabular}{p{2.4cm}p{5.5cm}}
Environment Models & Description  \\
\cline{1-2}\\[-1.0em]
Exploration tree ${\cal K}_j$ & Topological map of the environment: it stores the history of view configurations $\{\bm{q}^k_j\}$ in the form of a tree.\\
Volumetric map ${\cal H}_j$ & Representation of the explored region ${\cal E}^k$: it is used to compute the information gains of candidate view configurations.\\
Point cloud map ${\cal M}_j$ & Representation of the detected surfaces of the environment: it is used by the path planner to compute safe paths over segmented traversable regions and by the exploration agent to generate candidate view configurations. \\
\cline{1-2}
\end{tabular}
\end{center}
\end{table}

In this section, we introduce our approach. Fig.~\ref{Fig:ExplorationRobotModel} presents the main components of an exploration robot.
This interacts with its environment through observations and actions, where an observation consists of a vector of sensor measurements and an action corresponds to a robot actuator command. Team messages are exchanged with teammates over a network for sharing knowledge and decisions in order to attain team collaboration.  

Decision-making is achieved by the exploration agent and the path planner, basing on the available information stored in the environment model and the team model. In particular, the \emph{environment model} consists of a topological map, aka exploration tree, and two different metric maps: the volumetric map and the point cloud map.
The \emph{team model} represents the robot belief about the current plans of teammates (goals and planned paths).

The main components of the exploration robots are introduced in the following subsections. Table~\ref{Tab:EnvModels} reports their symbol along with a short description.

\begin{table*}[!ht]
\begin{center} 
\caption{Table of broadcast messages.}
\label{Tab:BroadcastMessages}
\setlength{\tabcolsep}{1em}
\renewcommand{\arraystretch}{1.1}
\begin{tabular}{p{2.7cm}|p{7.5cm}|p{6cm}}
Broadcast message & Content description  & Affected data in the recipient robot $h$ \\
\cline{1-3}
$\langle j, reached, \bm{g} \rangle$ & robot $j$ has reached goal position $\bm{g}$  & $j$-th tuple $(\bm{g}_j,\bm{\tau}_j,c_j)$ in team model ${\cal T}_h$ is reset\\
$\langle j, planned, \bm{g} \rangle$ & robot $j$ has planned node $\bm{g}$ as goal & $j$-th tuple $(\bm{g}_j,\bm{\tau}_j,c_j)$ in team model ${\cal T}_h$ is reset\\
$\langle j, selected, \bm{g}, \bm{\tau}, c \rangle$ & robot $j$ has selected node $\bm{g}$ as goal, $\bm{\tau}$ is the planned path and $c$ the corresponding path length  & $j$-th tuple in team model ${\cal T}_h$ is filled with $(\bm{g}, \bm{\tau}, c)$\\
$\langle j, aborted, \bm{g} \rangle$ & robot $j$ has aborted goal node $\bm{g}$  & $j$-th tuple $(\bm{g}_j,\bm{\tau}_j,c_j)$ in team model ${\cal T}_h$  is reset\\
$\langle j, position, \bm{p} \rangle$ & robot $j$ shares its current position $\bm{p}$  & $\bm{p}_j$ is set to $\bm{p}$ in the $j$-th tuple of ${\cal T}_h$ \\
$\langle j, \emph{scan}, \bm{v}, \bm{q}\rangle$ & robot $j$ has acquired new 3D scan data $\bm{v}$ at $\bm{q}$ &  3D scan data is integrated in volumetric map ${\cal H}_h$\\
$\langle j, \emph{tree}, {\cal K}_j \rangle$ & robot $j$ shares its  exploration tree ${\cal K}_j $ & ${\cal K}_h$ is compared with ${\cal K}_j$ (cfr. Algorithm~\ref{Alg:MapOverlapCheckAndSync}) to check if a map synchronization with robot $j$ is needed\\
\cline{1-3}
\end{tabular}
\end{center}
\end{table*}

\subsection{Exploration Tree and Exploration Agent}\label{Sect:ExplorationTree}

During the exploration process (see Sect.~\ref{Sect:ExplorationStrategies}), an \emph{exploration tree} ${\cal K}_j$ is built by robot $j$. A node of ${\cal K}_j$ is referred to as \emph{view node} and represents a view configuration. An edge between two view nodes corresponds to a safe path joining them. ${\cal K}_j$ is rooted at $\bm{q}_j^0$. At each forwarding step, a new node corresponding to  $\bm{q}_j^{k+1}$ and a new edge representing a path from $\bm{q}_j^{k}$  to $\bm{q}_j^{k+1}$  are added in ${\cal K}_j$.

We consider an exploration tree as a topological map representation of the explored environment. Each node of the tree represents an explored local region contained within a ball of radius $R_s$ centred at the associated view configuration.

The \emph{exploration agent} is responsible of online generating an exploration plan (cfr. Sect.~\ref{Sect:ExplorationStrategies}) according to the defined exploration objective (cfr. Sect.~\ref{Sect:ExplorationStrategies}). 

\subsection{Point Cloud Map and Path-Planning}\label{Sect:OverviewMappingTravPathPlanning}

Robot $j$ incrementally builds a 3D point cloud map ${\cal M}_j$ as a metric representation of the detected environment surfaces $ \bigcup\limits_{j = 1}^m \bigcup\limits_{i = 0}^k {\cal B}_j(\bm{q}_j^i)$.
The points of map ${\cal M}_j$ are segmented and partitioned into two main sets: \emph{obstacles} points ${\cal M}^{obs}_j$ and \emph{traversable} points ${\cal M}^{trav}_j$ (which can be ``back-projected" to a valid and safe robot configuration~\cite{fre2018patrolling}). Each point of  ${\cal M}_j$ is associated to a \emph{multi-robot traversability cost} function ${trav:\mathbb{R}^3 \to \mathbb{R}}$ \cite{fre2018patrolling}. 
In this regard, ${\cal M}_j$ and the traversability cost are used to associate a \emph{navigation cost} $J(\bm{\tau})$ to each safe path $\bm{\tau}$~\cite{fre2018patrolling}.

Given the current robot position $\bm{p}_r \in \mathbb{R}^3$ and a goal position $\bm{p}_g \in {\cal S}$, the \emph{path planner} computes the safe path $\bm{\tau}*$ which minimizes the navigation cost $J(\bm{\tau})$ and connect $\bm{p}_r$ with $\bm{p}_g$. The path planner reports a failure if a safe path connecting $\bm{p}_r$ with $\bm{p}_g$ is not found. See~\cite{fre2018patrolling} for further details. 

In this work, for simplicity, in order to reach a desired view configuration $\bm{q} = [\bm{p},\bm{\phi}]^T \in {\cal C}$, a robot first plans and follows a path towards $\bm{p}$, then it turns on itself on ${\cal S}$ so as to assume the closest orientation to $\bm{\phi}$.

\subsection{Volumetric Map and Information Gain}\label{Sect:VolumetricMapping}

A volumetric map ${\cal H}_j$ is incrementally built by robot $j$ to represent the explored region ${\cal E}^k$ and associate an information gain to each safe configuration. Specifically, ${\cal H}_j$ is a probabilistic occupancy gridmap, stored in the form of an Octomap~\cite{Hornung-2013}. This partitions the environment in \emph{free}, \emph{occupied} and \emph{unknown} cells with a pre-fixed resolution. Such a model allows to appropriately model an environment populated by low-dynamic objects~\cite{walcott2012dynamic}.

The information gain is defined and computed as follows.  At step $k$, the boundary of the explored region $\partial {\cal
E}^k$ is the union of two \emph{disjoint} sets:

\begin{itemize}

\item the \emph{obstacle boundary} $\partial {\cal E}^k_{\rm obs}$, i.e., the
part of $\partial {\cal E}$ which consists of detected obstacle
surfaces;

\item the \emph{free boundary} $\partial {\cal E}^k_{\rm
free}$, i.e., the complement of $\partial {\cal E}^k_{\rm obs}$,
which leads to potentially explorable areas.

\end{itemize}
One has $\partial {\cal E}^k_{\rm obs} = \bigcup_{i = 0}^k {\cal
B}(\bm{q}^i)$ and $\partial {\cal E}^k_{\rm free}=\partial {\cal E}^k
\setminus \partial {\cal E}^k_{\rm obs}$.

Let ${\cal V}(\bm{q},k)$ be the \emph{simulated view}, i.e., the view
which would be acquired from $\bm{q}$ if the obstacle boundary were
$\partial {\cal E}^k_{\rm obs}$. The information gain $I(\bm{q},k)$ is
a measure of the set of unexplored points lying in
${\cal V}(\bm{q} ,k)$. We compute $I(\bm{q},k)$ as the total volume of the unknown cells of ${\cal H}_j$ lying in
${\cal V}(\bm{q} ,k)$, following the approaches in \cite{KrGuWa:96,BaLa:00}.

\begin{figure*}
\removelatexerror
\begin{algorithm}[H]
{\small
 \DontPrintSemicolon
 \nonl \textbf{MapOverlapCheckAndSync}($\langle j, \emph{tree}, {\cal K}_j \rangle$)\;
\nonl~\hspace{-0.5cm}\codecomment{robot $h$ checks map overlap with robot $j$; robot $h$ compares received ${\cal K}_j$ with ${\cal K}_h$ and sends robot $j$ missing parts}\;
$\{^j\bm{q}_k\}_{k=1}^{l_j} \leftarrow {\cal K}_j$ \codecomment{extract view configurations from ${\cal K}_j$; $l_j$ is the number of extracted configurations}\;
$\{^h\bm{q}_k\}_{k=1}^{l_h} \leftarrow {\cal K}_h$ \codecomment{extract view configurations from ${\cal K}_h$; $l_h$ is the number of extracted configurations}\;
\nonl~\hspace{-0.5cm}\codecomment{iterate over extracted configurations of ${\cal K}_h$}\;
    \For{ $r=1$ {\rm \textbf{to}} $l_h$} 
    {
    \nonl~\hspace{-0.1cm}\codecomment{given $^h\bm{q}_r$ of ${\cal K}_h$, find its closest configuration in ${\cal K}_j$ and the corresponding Euclidean distance}\;
    $(^j\bm{q}^*, d^*) \leftarrow \text{arg}\underset{k \in \{1,...,l_j\}}{\text{min}}~\text{dist}_{3D}(^h\bm{q}_r, ^j\bm{q}_k)$  \codecomment{$\text{dist}_{3D}(\bm{q}_1,\bm{q}_2)$ is the 3D Euclidean distance $\vert \bm{q}_1^{x,y,z} - \bm{q}_2^{x,y,z} \vert $}\;
   \If{ $d^* > R_s$}
   {
	 send robot $j$ scan at $^h\bm{q}_r$ \codecomment{robot $j$ is missing the scan acquired at $^h\bm{q}_r$}\;
   }
    }
 \textbf{return} \emph{false};
}
\caption{MapOverlapCheckAndSync}
\label{Alg:MapOverlapCheckAndSync}
\end{algorithm}
\end{figure*}

\subsection{Network Model}\label{Sect:NetworkModel}

Each robot can broadcast messages in order to share knowledge, decisions and achievements with teammates. We assume each of these messages can be lost during its transmission with a given probability $P_c$. 

Different types of broadcast messages are used by the robots to convey various information and attain coordination. In this process, the identification number (ID) of the emitting robot is included in the heading of any broadcast message. In particular, a \emph{coordination broadcast message} is emitted by a robot in order to inform teammates when it reaches a goal node (\emph{reached}), planned a goal node (\emph{planned}), selected a goal node (\emph{selected}), abort a goal node (\emph{aborted}), acquires new laser data  (\emph{scan}). Additionally, a message \emph{tree} is broadcast in order to enforce the synchronization of the volumetric maps and point cloud maps amongst teammates (see Sect.~\ref{Sect:SharedKnowledge}). 
Table~\ref{Tab:BroadcastMessages} summarizes the used broadcast messages along with the conveyed information/data. The general message format is $\langle robot\_id, message\_type, data \rangle$. 

\subsection{Shared Knowledge Representation and Update}\label{Sect:SharedKnowledge}

Each robot of the team stores and updates its individual representation of the world state. In particular, robot $j$ incrementally builds its individual point cloud map ${\cal M}_j$ by using the acquired scans (see~\cite{fre2018patrolling}). 
This allows the path planner to safely take into account explored terrain extensions and possible environment changes. 

At the same time, robot $j$ updates its volumetric map ${\cal H}_j$  by integrating its acquired 3D scans and the \emph{scan} messages received from teammates. As mentioned above, some of the scan messages can be lost and ${\cal H}_j$ may only partially represent the actual explored region ${\cal E}^k$. 

In order to mitigate this problem, each robot continuously broadcasts a \emph{tree} message at a fixed frequency $1/T_M$. In particular, a recipient robot $h$ uses a \emph{tree} message from robot $j$ to estimate if the map ${\cal H}_h$ sufficiently overlaps with ${\cal H}_j$ or it missed the integration of some \emph{scan} message data. A pseudocode description of this procedure is reported in Algorithm~\ref{Alg:MapOverlapCheckAndSync}: this verifies if the Euclidean projections of ${\cal K}_j$ and ${\cal K}_h$ sufficiently overlap each other. If a sufficient overlap is not verified, a synchronization and merge procedure is triggered between the maps of the two robots $i$ and $j$ by requesting the missing scans (line 4, Algorithm~\ref{Alg:MapOverlapCheckAndSync}).

The above information sharing mechanism allows to implement a \emph{shared information gain}, i.e. each robot computes the information gain on the basis of a distributed shared knowledge. This enforces team cooperation by minimizing unnecessary actions such as  exploring regions already visited by teammates.

\subsection{Team Model}\label{Sect:TeamModel}

In order to cooperate with its teammates and manage conflicts, robot $j$ maintains an internal representation of the \emph{planning state of the team} by using a dedicated table:
\begin{equation}
{\cal T}_j = \langle(\bm{p}_1,\bm{g}_1,\bm{\tau}_1,c_1,t_1),...,(\bm{p}_m,\bm{g}_m,\bm{\tau}_m,c_m,t_m)\rangle.
\end{equation} 
This stores for each robot $h$: its current position $\bm{p}_h \in \mathbb{R}^3$, its selected goal position $\bm{g}_h \in {\cal S}$, the last computed safe path $\bm{\tau}_h$ to $\bm{g}_h$, the associated travel cost-to-go $c_h \in \mathbb{R}^+$ (i.e. the length of $\bm{\tau}_h$), and the timestamp $t_h \in T$ of the last message used to update any portion of $(\bm{p}_h,\bm{g}_h, \bm{\tau}_h,c_h)$. The table ${\cal T}_j$ is updated by using \emph{reached}, \emph{planned}, \emph{selected}, \emph{aborted} and \emph{position} messages. In particular, \emph{reached}, \emph{planned} and \emph{aborted} messages received from robot $h$ are used to reset the tuple $(\bm{g}_h,\bm{\tau}_h,c_h)$ to empty (i.e. no information available). A \emph{selected} message sets the tuple $(\bm{g}_h,\bm{\tau}_h,c_h,t_h)$. Furthermore, a \emph{position} message from robot $h$ is used to update its position estimate in ${\cal T}_j$. 

An \emph{expiration time} $\delta t_e$ is used to clean ${\cal T}_j$ of old and invalid information. In fact, part of the information stored in ${\cal T}_j$ may refer to robots that underwent critical failures or whose connections have been down for a while. 
A tuple $(\bm{p}_h,\bm{g}_h, c_h,\bm{\tau}_h,t_h)$ is reset to zero if $(t - t_h) > \delta t_e$, where $t \in T$ denotes the current time.

\section{Two-Level Coordination Strategy}\label{Sect:TwoLevelCoordinationStrategy}

This section first introduces the notions of topological and metric conflicts and then presents our two-level coordination strategy (see Sect.~\ref{Sect:TwoLevelCoordinationStrategy}). In this context, \emph{coordination} (avoid conflicts) and \emph{cooperation}
(avoid inefficient actions) are crucial team objectives.

\subsection{Topological and Metric conflicts}\label{Sect:TopologicalMetricConflicts}

A \emph{topological conflict} between two robots $i$ and $j$ is defined with respect to their exploration trees ${\cal K}_i$ and ${\cal K}_j$. 
Specifically, a \emph{node conflict} occurs when robot $i$ and $j$ attempt to add a new view node in the same spatial region at the same time, respectively $n_i$ in ${\cal K}_i$ and $n_j$ in ${\cal K}_j$.
We identify this situation when the corresponding positions $\bm{p}(n^i_g) \in \mathbb{R}^3$ and $\bm{p}(n^j_g) \in \mathbb{R}^3$ are closer than a certain distance $D_g \leq R_s$. 

On the other hand, metric conflicts are directly defined in the 3D Euclidean space where two robots are referred to be in \emph{interference} if their centres are closer than a pre-fixed \emph{safety distance} $D_s$.  It must hold $D_s \geq 2 R_b$, where $R_b$ is the common bounding radius of robots, i.e. the radius of their minimal bounding sphere.
A \emph{metric conflict} occurs between two robots if they are in interference or if their planned paths may bring them in interference\footnote{That is, the distance between the closest pair of points of the two planned paths is smaller than $D_s$. 
}.

It is worth noting that topological conflicts may not correspond to metric conflicts. In our framework, a node may represent a large region that could be visited by two or more robots at the same time. 

\begin{figure}[t]
\centering
\includegraphics[width=\columnwidth]{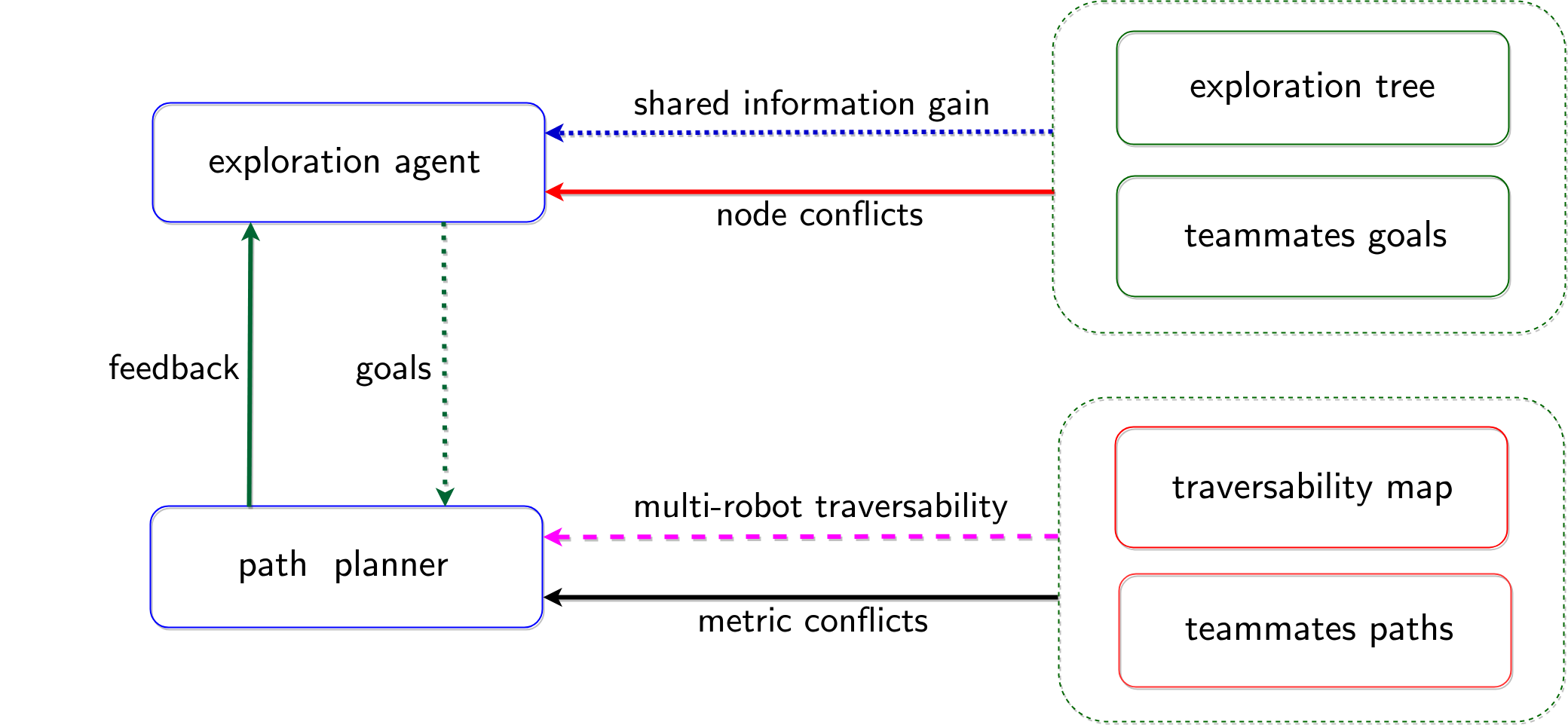}
\caption{The two-level strategy implemented on each robot.}
\label{Fig:TwoLevelStrategy}
\end{figure}

\subsection{Coordination Strategy}\label{Sect:TwoLevelCoordinationStrategy}

Our exploration strategy is distributed and supported on two levels: topological and metric (see Figure~\ref{Fig:TwoLevelStrategy}). 

The exploration agent acts on the \emph{topological} strategy level by suitably planning and adding a new view node $n_g$ (corresponding to a view configuration) on its exploration tree. \emph{Cooperation} is attained by planning $n_g$ on the basis of the shared information gain (see Sect.~\ref{Sect:SharedKnowledge}).

The path planner acts on the \emph{metric} strategy level (see Figure~\ref{Fig:TwoLevelStrategy}) by computing the best safe path from the current robot position to the goal position $\bm{p}(n_g)$ using its individual traversable map (see~\cite{fre2018patrolling}).

The exploration agent guarantees \emph{topological coordination} by continuously monitoring and negotiating possibly incoming node conflicts (see Sect.~\ref{Sect:ExplorationAgent}). 
In case two or more robots plan view configurations close to each other (node conflict), the robot with the smaller path length actually goes, while the other robots stop and re-plan a new view node. 

The path planner guarantees \emph{metric coordination} by applying a multi-robot traversability function. This induces prioritized path planning, in which robots negotiate metric conflicts by preventing their planned paths from intersecting (see~\cite{fre2018patrolling}).

The continuous interaction between the exploration agent and the path planner plays a crucial role. The exploration agent assigns desired \emph{goals} to the path planner. The latter in turn continuously replies with a \emph{feedback} informing the former about its state and computations. 
In particular, when the robot moves towards $\bm{p}(n_g)$, the path planner continuously re-plans the best traversable path until the robot reaches the assigned goal. During this process, 
if a safe path is not found, the path planner stops the robot, informs the exploration agent of a \emph{path planning failure}, and then the exploration agent re-plans a new view node. On the other hand, every time the path planner computes a new safe path, its length is used by the exploration agent to resolve possible node conflicts.

In our view, the two-way strategy approach allows to reduce interferences and manage possible deadlocks. 
In fact, while the exploration agent focuses on the most important exploration aspects (shared information gain maximization and node conflicts resolution), the path planner takes care of possible incoming metric conflicts.
Moreover, where the path planner strategy may fail alone in arbitrating challenging conflicts, the exploration agent intervenes and reassigns tasks to better redistribute robots over the environment. These combined strategies minimize interferences by explicitly controlling node conflicts and planning on the multi-robot traversable map.


\section{The Exploration Agent}\label{Sect:ExplorationAgent}


\begin{figure*}
\removelatexerror
\begin{algorithm}[H]\label{Alg:ExplorationAgent}
{\small
 \DontPrintSemicolon
 \nonl \textbf{ExplorationAgent}(\emph{robot\_id} $j$)\;
  \emph{current\_node} $\leftarrow$ new node at current robot configuration \codecomment{init current node, it represents $\bm{q}^k$}\;
  ${\cal K}_j$.\emph{root} $\leftarrow$ \emph{current\_node} \codecomment{init exploration tree ${\cal K}_j$, set its root node}\;  
  \emph{goal} $\leftarrow \emptyset$ \codecomment{init goal, it represents the next view node $\bm{q}^{k+1}$}\;  
  \emph{is\_goal\_reached}~$\leftarrow$ \emph{true}\;
 \While{true}
 {
 Update() \codecomment{update data structures and boolean variables}\;
 \eIf{ {is\_goal\_reached} } 
 {	
 ${\cal K}_j$ $\leftarrow$ add \emph{goal} as new node and create a new edge between \emph{current\_node}  and \emph{goal}\;
 \emph{current\_node} $\leftarrow$ \emph{goal} \codecomment{set goal as current node in the exploration tree}\; 
 broadcast \emph{goal} is \emph{reached} \;
 \emph{goal} $\leftarrow$~PlanNextBestView(\emph{current\_node}) \codecomment{plan next view node}\;
 broadcast \emph{goal} is \emph{planned} \;
 send \emph{goal} to path planner as goal\;   
 }
 {	
 	\eIf{ {(is\_path\_planning\_failure} {\rm \textbf{or}} {is\_node\_conflict)} }
 	{
 	send \emph{goal-abort} to path planner\;
 	broadcast \emph{goal} is \emph{aborted} \;  
 	\emph{goal} $\leftarrow$~PlanNextBestView(\emph{current\_node})\codecomment{plan next view node}\;
 	broadcast \emph{goal} is \emph{planned}\;
 	send \emph{goal} to path planner as goal\; 	
 	} 	
 	{
    broadcast \emph{goal} is \emph{selected} \codecomment{emit the select message at each step for robustness}\;
 	sleep for $T_{sleep}$\;
 	}
 }
 }
 \caption{ExplorationAgent (in robot $j$)}
}
\end{algorithm}
\smallskip

\removelatexerror
\begin{algorithm}[H]\label{Alg:Update}
{\small
 \DontPrintSemicolon
 \nonl \textbf{Update}()\;
    update volumetric map ${\cal H}_j$ by using received \emph{scan} messages and last acquired scan\;
    update point cloud map ${\cal M}_j$ by using last acquired scan\;    
    update team model ${\cal T}_j$ by using \emph{reached}, \emph{planned}, \emph{selected}, \emph{aborted} messages\;
    \emph{is\_node\_conflict}~$\leftarrow$~ check if the goal of another robot is closer than $D_g$ to current \emph{goal}\;
    \emph{is\_goal\_reached}~$\leftarrow$~check if current goal node has been reached \; 
    \emph{is\_path\_planning\_failure}~$\leftarrow$~check if path planner failed to compute a path to current goal node\;
 \caption{Update (in robot $j$)}
 }
\end{algorithm}
\smallskip

\removelatexerror
\begin{algorithm}[H]\label{Alg:SelectNexNode}
{\small
 \DontPrintSemicolon
 \nonl \textbf{PlanNextBestView}(\emph{current\_node})\;
    \emph{next\_node} $\leftarrow \emptyset$ \codecomment{init next view node}\;
    \emph{box} $\leftarrow \emptyset$ \codecomment{init search bounding box where to constrain admissible set construction}\;    
    \emph{bounding\_boxes} $\leftarrow \{\emph{box}_1,\emph{box}_2,...,\emph{box}_l\}$ \codecomment{pre-defined sequence of increasing bounding boxes}\;       
    \ForEach{box {\rm \textbf{in}} bounding\_boxes }{%
	 	${\cal D}$~$\leftarrow$~BuildSearchTree(\emph{current\_node}, \emph{box})\codecomment{build the admissible set as a tree contained in bounding box}\;
 		$\bm{q}^*$ $\leftarrow$~select node with maximum utility in ${\cal D}$ \codecomment{receding-horizon optimization, equivalent to select "best" branch}\;
 		\If{ $U(\bm{q}^*) > U_{min}$}
 		{
 		\textit{branch} $\leftarrow $ extract the branch of ${\cal D}$ containing $\bm{q}^*$ \codecomment{"best" branch extraction}\;
 		\emph{next\_node} $\leftarrow$ select the node of \emph{branch} at prefixed graph-distance $\delta_D$ from  ${\cal D}$.\emph{root} \codecomment{forwarding step}\;
 		\textbf{break} \codecomment{exit from foreach loop} \;  
 		} 	

      } 
 	\If{ next\_node $== \emptyset$}
 	{
 		\emph{next\_node} $\leftarrow$~BackTrackingStrategy() \codecomment{backtracking step}\; 		
 	} 	   
 	\textbf{return} \emph{next\_node}\;
 \caption{PlanNextBestView (in robot $j$)}
}
\end{algorithm}
\smallskip

\removelatexerror
\begin{algorithm}[H]\label{Alg:BuildAdmissibleSet}
{\small
 \DontPrintSemicolon
 \nonl \textbf{BuildSearchTree}(\emph{current\_node}, \emph{box})\;
   ${\cal D}$.\emph{root} $\leftarrow$ \emph{current\_node} \codecomment{init search tree ${\cal D}$, set its root node}\; 
 ${\cal D}$~$\leftarrow$ expand a sampling-based tree on the surface ${\cal M}^{trav}_j \cap \emph{box}$ \;
  compute information $I(\bm{q})$ and utility $U(\bm{q})$ for each sampled configuration $\bm{q}$ in ${\cal D}$\;
  \textbf{if} $I(\bm{q}) > I_{min}$ \textbf{then} add $\bm{q}$ to the frontier tree ${\cal Y}_j$\;
 \textbf{return} ${\cal D}$\;
 \caption{BuildSearchTree (in robot $j$)}
 }
\end{algorithm}
\smallskip

\removelatexerror
\begin{algorithm}[H]\label{Alg:BacktrackingStrategy}
{\small
 \DontPrintSemicolon
 \nonl \textbf{BacktrackingStrategy}(\emph{current\_node})\;
 update information gain of each frontier node in the frontier tree ${\cal Y}_j$\; 
 build frontier node clusters and compute the cluster centroids in ${\cal Y}_j$\;
 compute utility of each frontier cluster centroid \codecomment{according to Sect.~\ref{Sect:BacktrackingStrategy}}\;
 \textbf{return} best frontier cluster centroid\;
 \caption{BacktrackingStrategy (in robot $j$)}
 }
\end{algorithm}

\end{figure*}

In this section, we present in detail the exploration agent algorithm. A pseudocode description is reported in Algorithm~\ref{Alg:ExplorationAgent}. Some comments accompany the pseudocode  with the aim of rendering it self-explanatory. 
 
An exploration agent instance runs on each robot. It takes as input the robot ID $j$. A main loop supports the exploration algorithm (lines 5--26).
First, all the main data structures along with the boolean variables\footnote{We use an ``\emph{is\_}" prefix to denote boolean variables.} are updated (line 6). This update takes into account all the information received from teammates and recast the distributed knowledge. 
When the current goal is reached (line 7), it is added as a new node in ${\cal K}_j$. A new edge is created from the current node $\bm{q}^{k}$ to the new node $\bm{q}^{k+1}$. Next, a new node is planned and sent to the path planner as the next goal position (line 13). During this steps, the robot informs the team about its operations by using broadcast messages (line 10 and 12).

On the other hand, if the robot is still reaching the current goal node, lines 15--24 are executed. 
If either a path planner failure or a node conflict (see Section~\ref{Sect:TeamNodeConflict}) occurs (line 15) then the exploration agent sends a goal abort to the path planner and broadcasts its decision (lines 16--17). This triggers a new node selection (line 18). Otherwise (line 22), a selected message is emitted for sake of robustness (aiming at maximizing the probability the message is actually received by all teammates). Then, a sleep for a pre-fixed time $T_{sleep}$ allows the robot to continue its travel towards the planned goal (line 23). 

It is worth noting that the condition at line 15 of Algorithm~\ref{Alg:ExplorationAgent} implements continuous monitoring and allows robots to modify their plan while moving.

\subsection{Data Update}\label{Sect:Update}

The Update() function is summarized in Algorithm~\ref{Alg:Update}. This is in charge of refreshing the robot data structures presented in Sect.~\ref{Sect:Method}. Indeed, these structures are asynchronously updated by callbacks which are independently triggered by new teammates broadcast messages or path planner feedback messages.

Lines 1--3 of Algorithm~\ref{Alg:Update} represent the asynchronous updates of the volumetric map ${\cal H}_j$, the point cloud map ${\cal M}_j$ and of the team model ${\cal T}_j$. The remaining lines describe how the reported boolean variables are updated depending on the information stored in the team model and received from path planner feedback.

\subsection{Node Conflict Management}\label{Sect:TeamNodeConflict}

The concept of topological node conflict was defined in Section~\ref{Sect:TopologicalMetricConflicts}. During an exploration process, this occurs when (i) two robots select their goals at a distance closer than a pre-fixed value $D_g$ or (ii) when a robot selects its goal at distance closer than $D_g$ from a teammate position. The first case is referred to as \emph{goal-goal conflict}, while the latter is called \emph{goal-start conflict}.   

A robot checks for node conflicts by solely using the information stored in its individual team model. Robot $j$ can detect a node conflict with robot $i$ only if, in its team model ${\cal T}_j$, the tuple $(\bm{p}_i,\bm{g}_i, c_i,\bm{\tau}_i,t_i)$ is valid (see Sect.~\ref{Sect:TeamModel}). 

In particular, robot $j$ detects a goal-goal conflict with robot $i$ if the following conditions are verified:
\begin{enumerate}
\itemsep0em 
\item a path connecting the goals $\bm{g}_j$ with $\bm{g}_i$ exists on the local traversability map of robot $j$, and its length is smaller than $D_g$.
\item $c_j$ is higher than $c_i$ in ${\cal T}_j$, or $j>i$ in the unlikely case the navigation costs are equal (robot priority by ID as fallback). 
\end{enumerate}
This entails that the priority is given to the robot with the smallest navigation cost. 

Robot $j$ detects a goal-start conflict with robot $i$ if a path connecting the goal $\bm{g}_j$ with the position $\bm{p}_i$ exists on the local traversability map of robot $j$, and its length is smaller than $D_g$.

When one of the two above conflict cases occurs, the boolean variable is\_node\_conflict is set to true by robot $j$ (line 3 of Alg.~\ref{Alg:Update}). As a consequence, its current goal is aborted and a new node is planned (lines 16--20 of Alg.~\ref{Alg:ExplorationAgent}). 
 
\begin{figure}[!ht]
\begin{center}   
     \subfloat[\label{teaser1}]{%
       \includegraphics[height=\columnwidth]{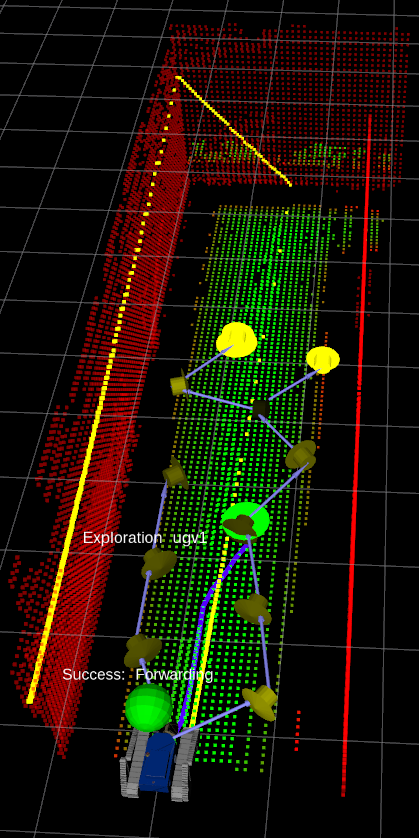}
     } \\         
     \subfloat[\label{teaser2}]{%
       \includegraphics[height=0.5\columnwidth]{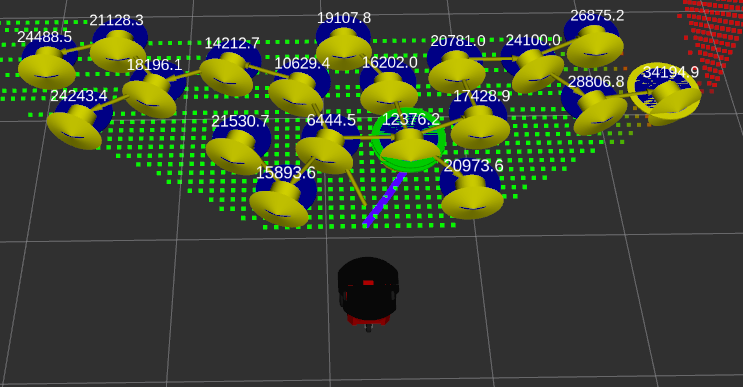}
     }            
     \caption{\textit{(a)} A robot, its point cloud map and the expanded search tree during an exploration step. Red points represent the segmented obstacles points ${\cal M}^{obs}_j$. Green points represent the traversable points ${\cal M}^{trav}_j$. The volumetric map is not shown. The small arrows represent candidate view configurations: the lighter the yellow, the higher the corresponding utility value. \textit{(b)} Another example with a pioneer robot and its seach tree. In both pictures, the nodes marked with a yellow disk represent the candidate view nodes with the highest utility (best branch) and the nodes marked in green represent the selected next view nodes at pre-fixed graph distance.}
     \label{Fig:SearchTreeConstruction} 
\end{center}
\end{figure}

\subsection{Next Best View Selection Strategy}

This section describes the PlanNextBestView() function, whose pseudocode is reported in Algorithm~\ref{Alg:SelectNexNode}. This function is responsible of generating an admissible set ${\cal D}$ and selecting the next best view configuration in ${\cal D}$ (forwarding strategy, cfr. Sect.~\ref{Sect:ExplorationStrategies}). 

In order to achieve efficiency and a quick response time, a windowed search strategy is adopted. In particular, in Algorithm~\ref{Alg:SelectNexNode}, the main loop (lines 4--12) guides the search of the next best view through an increasing sequence of bounding boxes $\{\emph{box}_i\}_{i=1}^l$, where $\emph{box}_i \subset \emph{box}_{i+1}$ (lines 4--9).

At each iteration (line 4, Algorithm~\ref{Alg:SelectNexNode}), an admissible set ${\cal D}$ is created in the current bounding box (line 5), in the form of a tree by using the BuildSearchTree() function. In the following, we refer to ${\cal D}$ as the \emph{search tree}. A utility $U(\bm{q})$ (see Sect.~\ref{Sect:SearchTreeConstruction}) and an information gain $I(\bm{q})$ (see Sect.~\ref{Sect:OverviewMappingTravPathPlanning}) are associated to each candidate configuration $\bm{q}$ in ${\cal D}$.  The next best view configuration  $\bm{q}^*$ is computed as the one maximizing the utility in ${\cal D}$ (line 7).

If the utility $U(\bm{q}^*)$ is greater than a minimum threshold $U_{min}$, then $\bm{q}^*$ is actually used for a forwading step (line 8--9). On the other hand, if a valid next view configuration is not found (line 13) then the robot performs a backtracking step (line 14, Algorithm~\ref{Alg:BacktrackingStrategy}, Sect.~\ref{Sect:BacktrackingStrategy}). 

The selection of the next view node is performed at line 9 of Algorithm~\ref{Alg:SelectNexNode}) along the best branch of the search tree~(see next section).

\subsection{Search Tree Construction}\label{Sect:SearchTreeConstruction}

The construction of the admissible set ${\cal D}$ (line 5, Algorithm~\ref{Alg:SelectNexNode}) is performed by the BuildSearchTree() function, whose pseudocode is introduced in Algorithm~\ref{Alg:BuildAdmissibleSet}. Each node $n_h$ of the search tree ${\cal D}$ represents a candidate view configuration $\bm{q}_h$, while an edge represents a safe path $\bm{\tau}$ in ${\cal C}$ between the two joined nodes (see Fig.~\ref{Fig:SearchTreeConstruction}).

The function BuildSearchTree() expands a sampling-based tree $\cal D$ on the set of traversable points ${\cal M}^{trav}_j \cap \emph{box}$. The root of $\cal D$ is initialized at the \emph{current\_node}. The tree is grown by using a breadth-first expansion over ${\cal M}^{trav}_j$ (line 2, Algorithm~\ref{Alg:BuildAdmissibleSet}). The use of the traversable points constrains the tree to automatically grow over a cloud of reachable candidate configurations.

Let $n_h$ denote a node in $\cal D$ and denote by $b$ its corresponding branch, i.e. the path in $\cal D$ which brings from the root to $n_h$. The utility of $n_h \in {\cal D}$ is computed as the total information gain which would be attained by a robot when moving towards $n_h$ along $b$. Specifically, the \emph{utility} $U(n_h)$ of a node $n_h \in {\cal D}$ with parent node $n_{h-1}$ is recursively defined as follows:
\begin{equation}\label{Eq:NodeUtility}
U(n_h) = U(n_{h-1}) + I(n_h) e^{-\lambda \sigma(n_h)},
\end{equation}
where $\sigma(n_h) = \sigma(n_{h-1}) + \|\bm{q}_h - \bm{q}_{h-1}\|$ represents the geometric length of the path connecting the root of $\cal D$ to the node $n_h$~\cite{bircher2016receding}. 

Such a definition of the utility function entails a receding-horizon optimization approach where, in fact, the ``best" branch (and not necessarily the best configuration) is selected (line 8, Algorithm~\ref{Alg:SelectNexNode}). Next, the robots moves along a pre-fixed number of edges along the selected branch.  

\subsection{Backtracking Strategy}\label{Sect:BacktrackingStrategy}

The backtracking strategy is achieved by building and updating the following data structures.

\begin{itemize}
\item \emph{Frontier nodes}: nodes whose information gain is greater than a prefixed threshold $I_{min}$ at the present time. During each BuildSearchTree() expansion, each new node of ${\cal D}$ that is a frontier node is added to the frontier tree. 
\item \emph{Frontier tree} ${\cal Y}_j$: it is incrementally expanded by attaching each new found frontier node to its nearest neighbor in the tree. To this aim, a k-d tree is efficiently used and maintained. Frontier nodes that are inserted in the frontier tree and subsequently become normal nodes (i.e. with $I \leq I_{min}$) are kept within the frontier tree to preserve its connectivity.  
\item \emph{Frontier clusters}: Euclidean clusters of close frontier nodes belonging to the frontier tree. These clusters are computed by using a pre-fixed clustering radius. Each cluster has an associated total information gain (the sum of the information gains of all the nodes in the cluster) and a corresponding representative centroid.
\end{itemize}

When a backtracking step is in order (see Algorithm~\ref{Alg:BacktrackingStrategy}), first the frontier clusters are built. Then, for each frontier cluster $j$, the navigation cost-to-go $c_j$ of its centroid (from the current robot position) and its total information gain $I_j$ are computed. Next, the utility of each cluster is computed as:
\begin{equation}
U_j = I_j e^{-\lambda c_j}
\end{equation} 
similarly to eq.~(\ref{Eq:NodeUtility}). Finally, the cluster centroid with the maximum utility is selected and planned as next best view node. 

While moving towards a selected backtracking node, the robot periodically runs the backtracking strategy (Algorithm~\ref{Alg:BacktrackingStrategy}) to confirm its last backtracking plan remains valid and informative along the way. This allows the robot to promptly take into account new covered regions (due to new asynchronous teammate scans) and avoid unnecessary travels.

\subsection{Prioritized Exploration}

Robot $j$ uses a \emph{priority queue} ${\cal P}_j$ in order to store a set of Point Of Interests (POI) with their associated priorities. The higher the priority the more important the POI in ${\cal P}_j$. At each exploration step, the most important point of interest is selected in ${\cal P}_j$ and used to bias the robot search for the next view configuration. 

In particular, the priority queue can be manually created by the end-user or can be autonomously generated by an object detection module. In this second case, the module automatically detects and classifies objects in the environment and assigns them priorities according to a pre-fixed class-priority table.  

Let $\bm{p}^*$ be the POI with the highest priority in ${\cal P}_j$. In BuildSearchTree() (see Algorithm~\ref{Alg:BuildAdmissibleSet}), a randomized A* expansion is biased towards $\bm{p}^*$. 
Specifically, at each step of such A* expansion, a coin toss with probability 0.5 decides if a heuristic function with input $\bm{p}^*$ is used or not to bias the breadth-first expansion. In this process, we use a simple Euclidean distance as heuristic function $h(\bm{p})$. The point $\bm{p}^*$ is deleted from ${\cal P}_j$ once the distance between the robot and $\bm{p}^*$ is smaller than the sensor perception range $R_s$.

\section{Coverage Task}\label{Sect:Coverage}

Coverage and exploration tasks have the same objective: robots have to cover the environment with sensory perceptions. But while coverage robots have a full prior knowledge of the environment ${\cal W}$ (which can be used to plan safe motions), exploration robots must discover ${\cal W}$ online and restrict their motions within the known terrain. 

A \emph{coverage plan} is a view plan $\pi_j$ (see Sect.~\ref{Sect:ExplorationTasks}) composed of view configurations $\bm{q}_j^k$ which are valid and reachable through a safe path from $\bm{q}^0_j$ (see Sect.~\ref{Sect:3DEnvModel}). Note that, given the a priori knowledge of the full environment, a coverage plan $\pi_j = \{\bm{q}_j^k\}$ includes view configurations which are not necessarily exploration-safe at each step $k$.
A \emph{team coverage strategy} $\Pi = \{\pi_1,...,\pi_m\}$ collects the coverage plans of the robots in the team. 

\noindent {\bf Coverage objective}. The robots must cooperatively plan a team coverage strategy $\Pi$ of minimum length $l$ such that ${\cal E}^l$ is coverage-maximized. In particular, ${\cal E}^l$ is \emph{coverage-maximized} 
at step $l$ if there is no robot $j$ which can plan a valid configuration $\bm{q}$ reachable through a safe path from $\bm{q}^0_j$ such that ${\cal E}^l \subset ({\cal E}^l \cup {\cal V}_j(\bm{q}))$.

The team of exploration robots can be used to perform a coverage task. To this aim, we provide them with a point cloud map ${\cal M}$ representing the full environment. Specifically, at $t_0$, each robot $j$ sets ${\cal M}_j = {\cal M}$ and ${\cal H}_j = \emptyset$. During the coverage task, the volumetric map ${\cal H}_j$ is grown and maintained as in a normal exploration process: it is incrementally built in order \emph{(i)} to represent the regions of the environment covered so far and \emph{(ii)} to compute the information gain of candidate view configurations. The map ${\cal M}_j$ is used by the path planner to plan safe paths all over the modeled environment terrain.

Clearly, such an approach does not take advantage of the full prior knowledge of the environment, which in principle allows to pre-compute an optimal team plan. Nonetheless, our framework can be conveniently used when the environment undergoes low-dynamic changes. In fact, our 3D mapping is capable of continuously integrating detected changes and our online decision-making correspondingly adapts robot behaviour to the new environment model.


\section{Results}\label{Sect:Results}
\begin{figure}[!t]
\begin{center}
     \subfloat[\label{pioneer_robot_conf}]{%
       \includegraphics[height=0.23\textwidth]{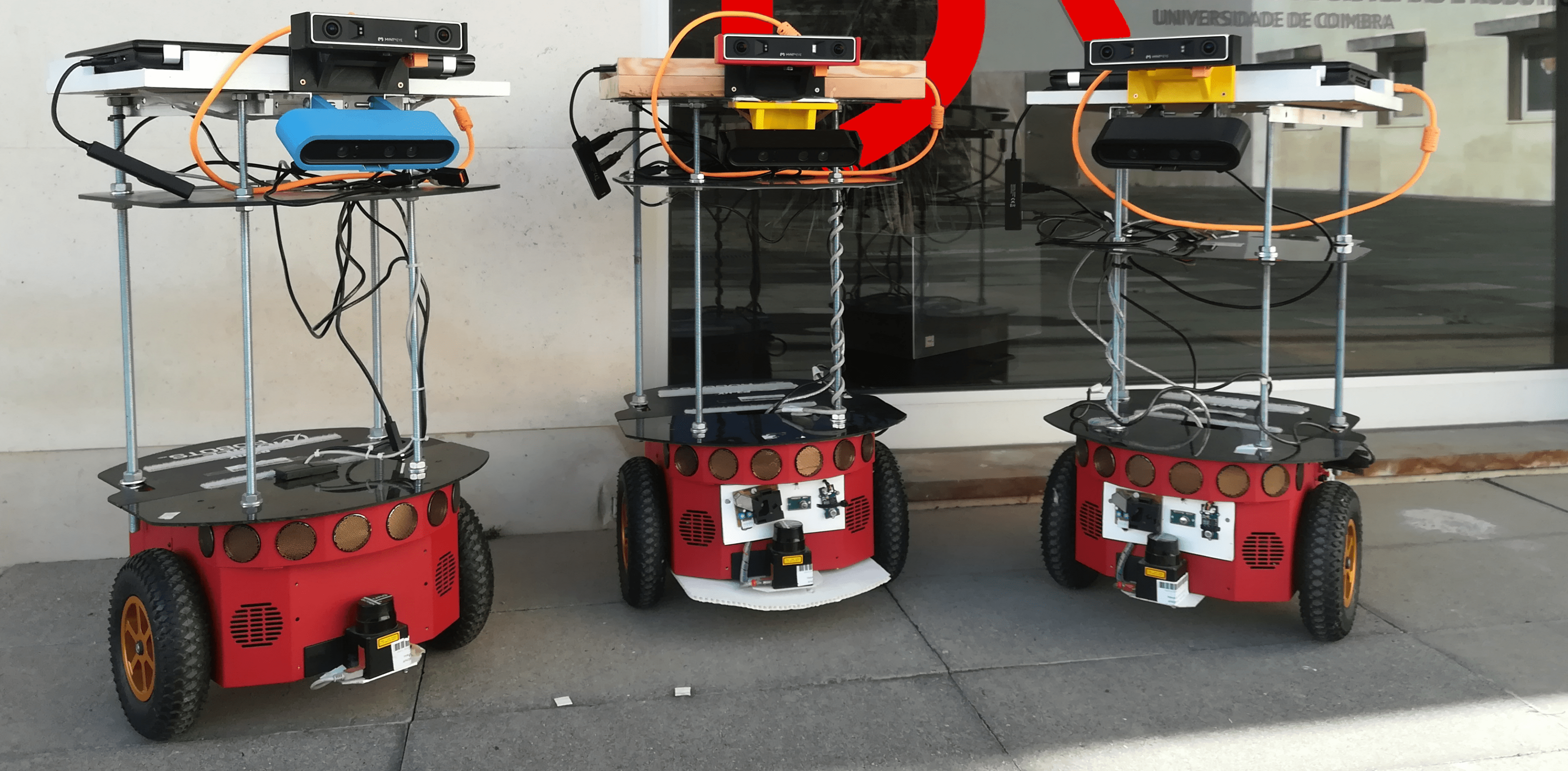}
     }
     \quad
     \subfloat[\label{pioneer_robot_fov}]{%
       \includegraphics[height=0.28\textwidth]{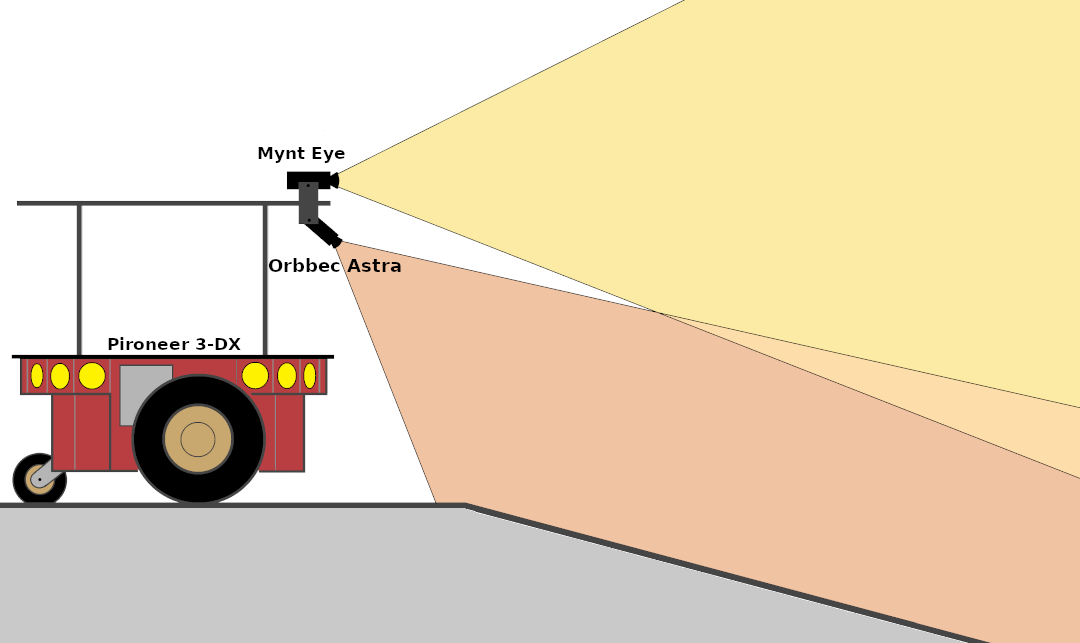}
     }      
     \caption{\textit{(a)} The team of 3 pioneer 3-DX robots used in our experiments at the University of Coimbra. The robots are equipped with two RGBD cameras: the Orbbec Astra S and the Mynt Eye S1030. \textit{(b)} A pioneer 3-DX sensor equipment together with the sensor FOVs.}
     \label{Fig:PioneerRobot} 
 \end{center}
 \end{figure}

\begin{figure}[!t]
\centering
	\includegraphics[width=\linewidth]{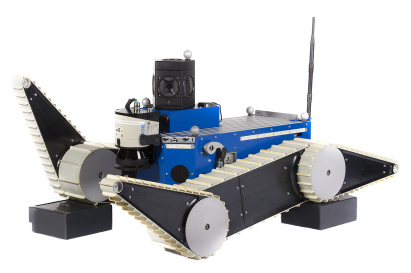}
    \caption{TRADR UGV equipped with multiple encoders, an IMU, and a rotating laser-scanner.}%
    \label{Fig:Robot}
\end{figure}

 \begin{figure*}
\begin{center}
     \subfloat[\label{pioneer_lab}]{%
       \includegraphics[height=0.225\textwidth]{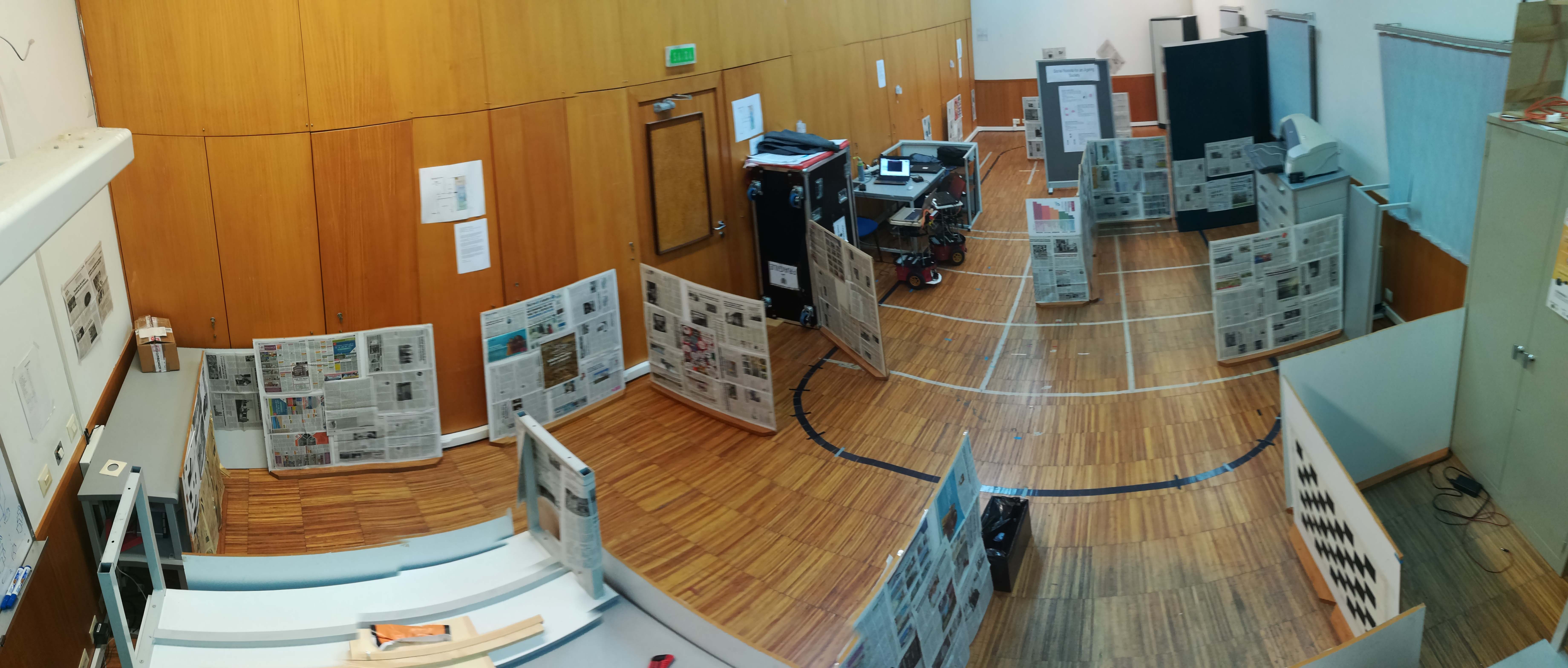}
     }
     \quad
     \subfloat[\label{pioneer_lab_map}]{%
       \includegraphics[height=0.225\textwidth]{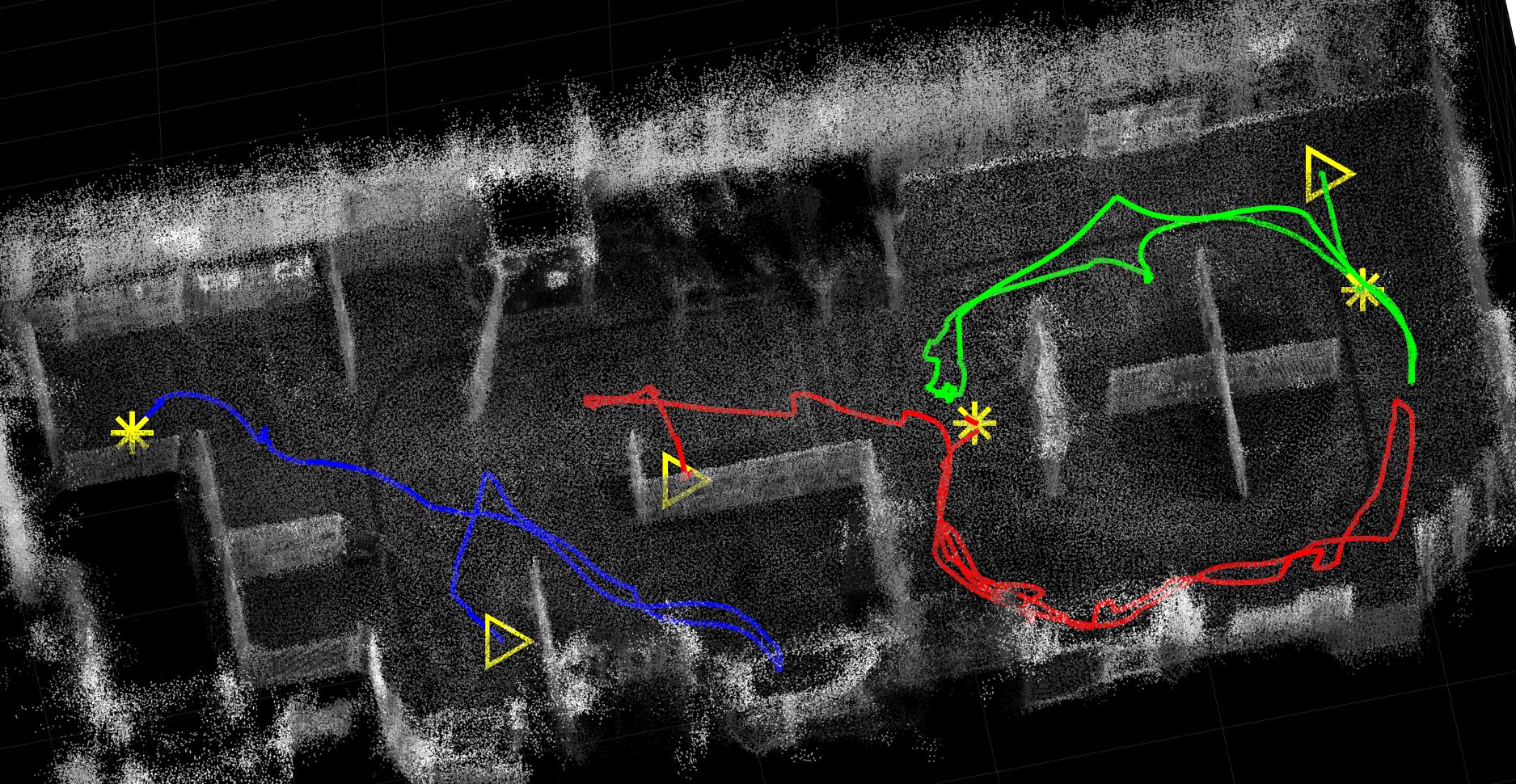}
     }      
     \caption{\textit{(a)}: One of the environments that was used for conducting experiments. \textit{(b)}: The map and robot traveled paths obtained after an exploration process with a team of 3 pioneer 3-DX robots.}
     \label{Fig:PioneerRobotExperiment} 
 \end{center}
 \end{figure*}
 
\begin{figure}[!t]
\begin{center}
\includegraphics[height=0.44\textwidth]{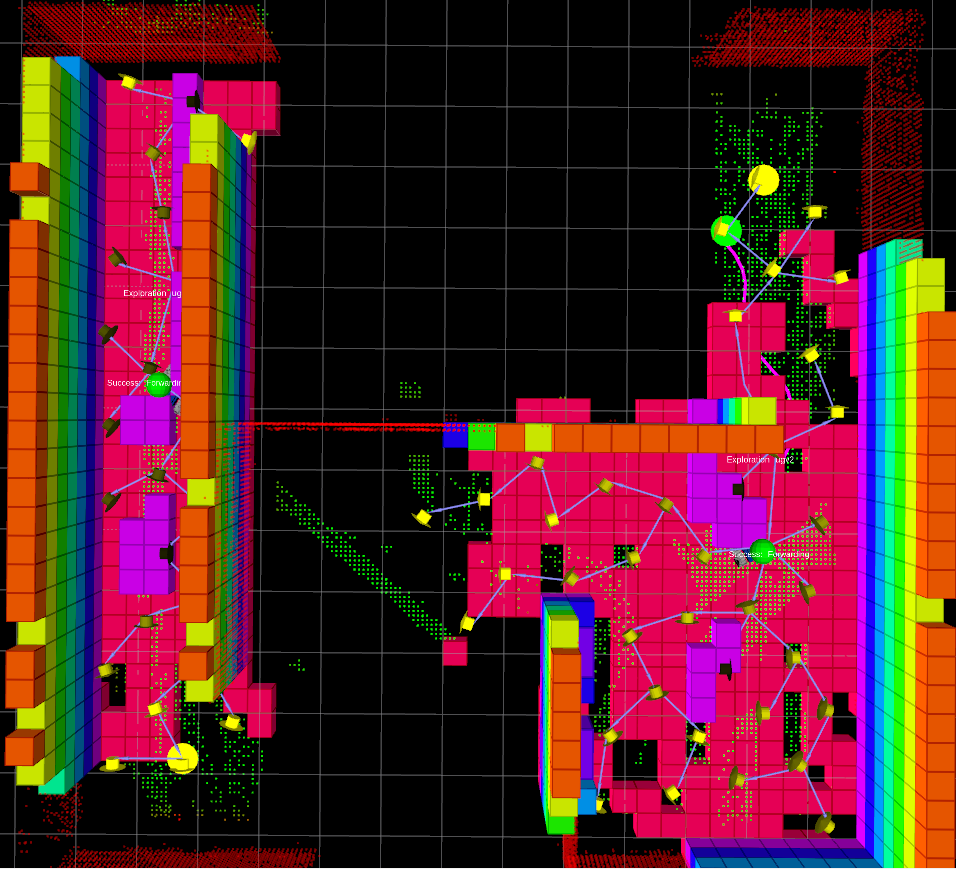}
\caption{A snapshot of a V-REP simulation while a team of two TRADR robots is exploring an environment. The two different volumetric maps of the robots are overlapped and shown in the picture. The robots start at different locations, knowing their initial mutual positions. As for the volumetric map, only the obstacle cells are shown. The search tree of each robot is also shown. Each search tree is expanded only on the traversable portion of the point cloud map.}
\label{fig:sim1} 
\end{center}
\end{figure}

In this section, we present experimental results we obtained with a ROS-based implementation of the proposed~method. 

Before conducting real-world experiments, we used V-REP as simulation framework~\cite{vrep}. V-REP allows to simulate laser range finders, depth cameras, odometry noise, and robot tracks with grousers to obtain realistic robot interactions with the terrain. Fig.~\ref{fig:sim1} shows a phase of a simulation test with an exploring team of two robots. A video showing a simulated exploration process with 3 robots is available at \href{https://www.youtube.com/watch?v=UDddKJck2yE}{www.youtube.com/watch?v=UDddKJck2yE}.

\subsection{Evaluations at the University of Coimbra}

An analysis of the proposed method via both simulation and real-world experiments was conducted in a Master Thesis project~\cite{Novo2021}. In this work, a team of Pioneer 3-DX robots was used to perform both simulations and real-world experiments (see Fig.~\ref{Fig:PioneerRobot}). In particular, the Pioneer robots were equipped with two RGBD cameras, the Orbbec Astra S and
the Mynt Eye S1030. 

Figure~\ref{Fig:PioneerRobotExperiment} shows one of the testing environments together with one of the obtained maps and the corresponding paths traveled by the robots.
A video showing some phases of the performed exploration experiments is available at \href{https://youtu.be/xwCnQtLC0RI}{youtu.be/xwCnQtLC0RI}.

\begin{figure*}
\begin{center}
     \subfloat[\label{subfig-1:exploration-detalinq}]{%
       \includegraphics[height=0.38\textwidth]{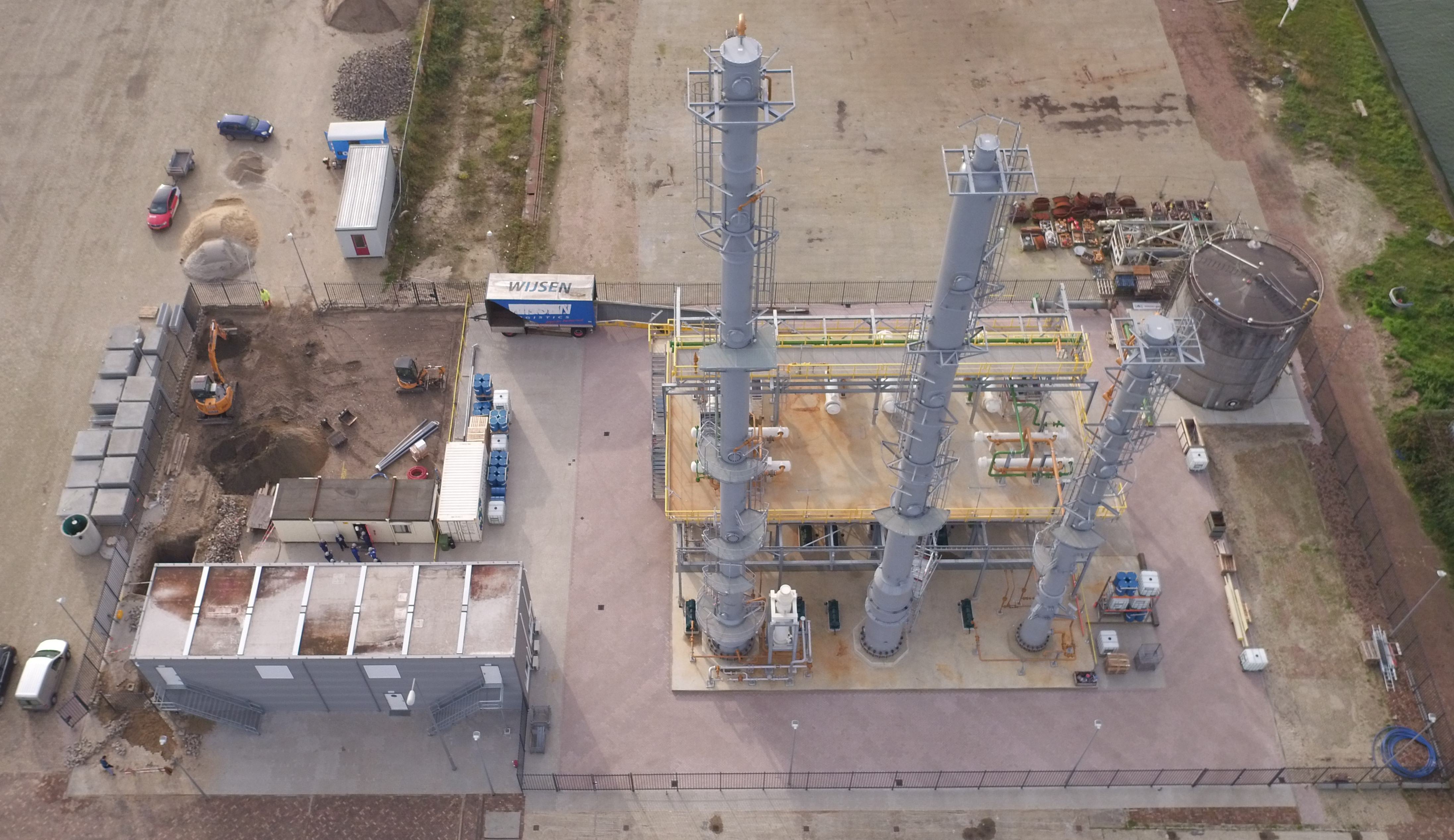}
     }
     \quad
     \subfloat[\label{subfig-2:fireman}]{%
       \includegraphics[height=0.38\textwidth]{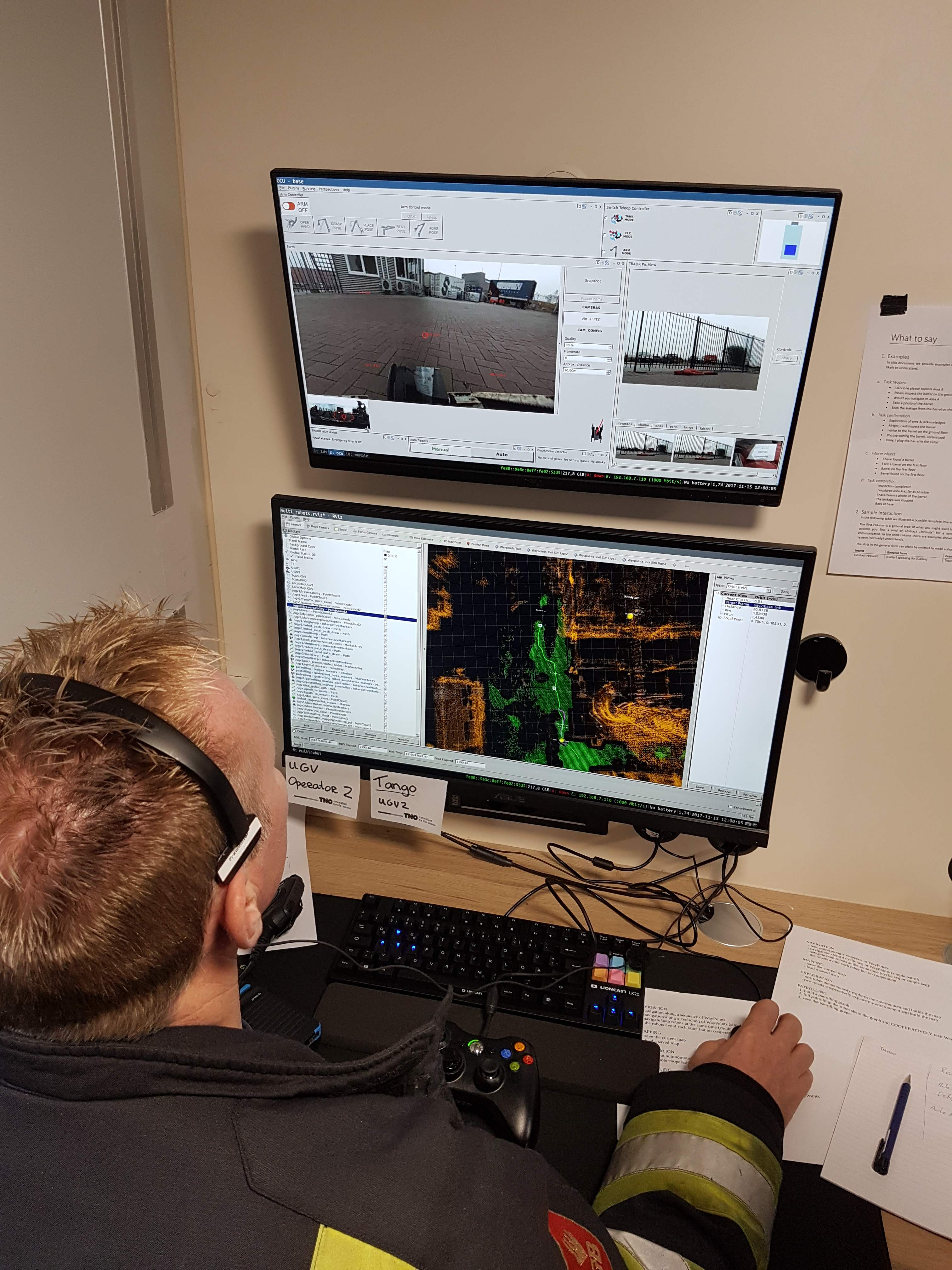}
     }
     \quad
     \subfloat[\label{subfig-2:map1}]{%
       \includegraphics[height=0.36\textwidth]{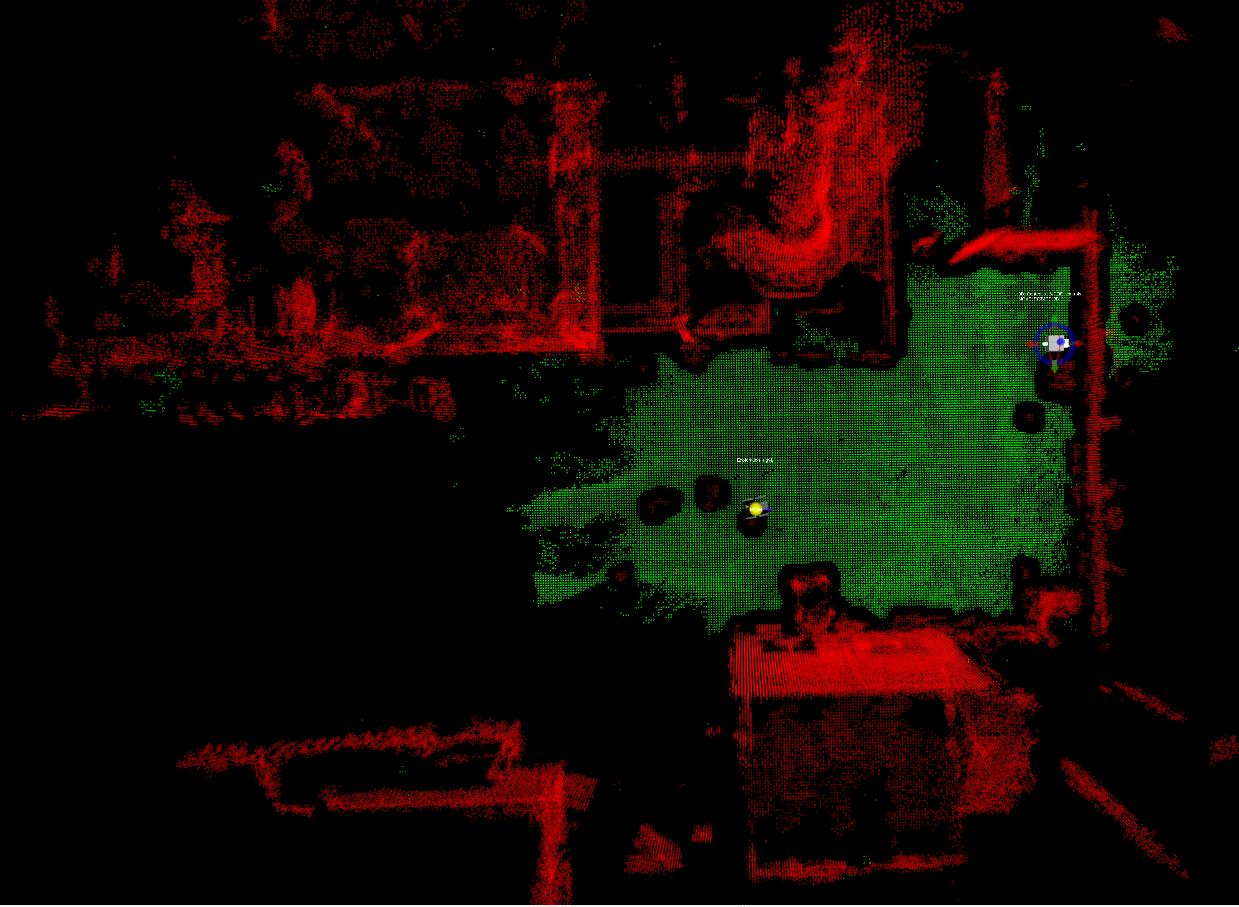}
     }        
     \quad
     \subfloat[\label{subfig-2:map2}]{%
       \includegraphics[height=0.36\textwidth]{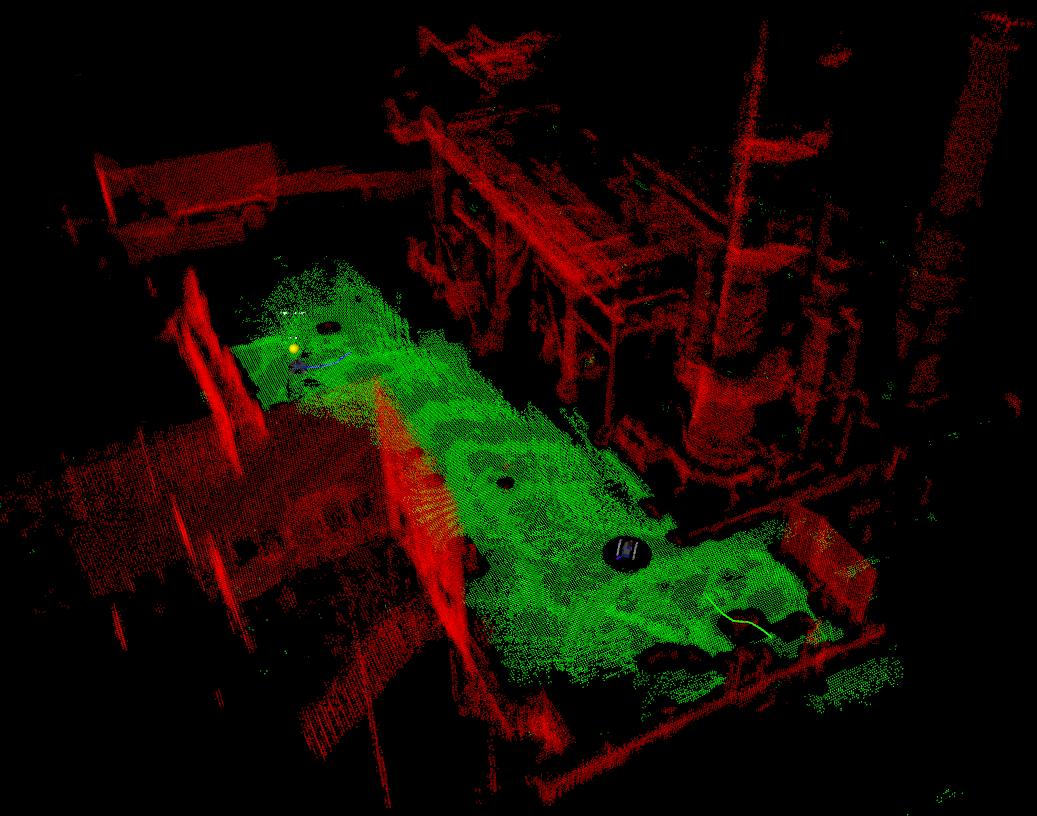}
     }       
     \caption{\textit{(a)}: the RDM training plant (Rotterdam) used for our first evaluations. \textit{(b)}: a fireman operating the GUI interface for monitoring path planning and exploration. \textit{(c)} and \textit{(d)}: two maps obtained in two different phases of a collaborative exploration process.}
     \label{fig:exploration-deltalinq1} 
 \end{center}
 \end{figure*}

\begin{figure}[!t]
\begin{center}
\includegraphics[width=0.48\textwidth]{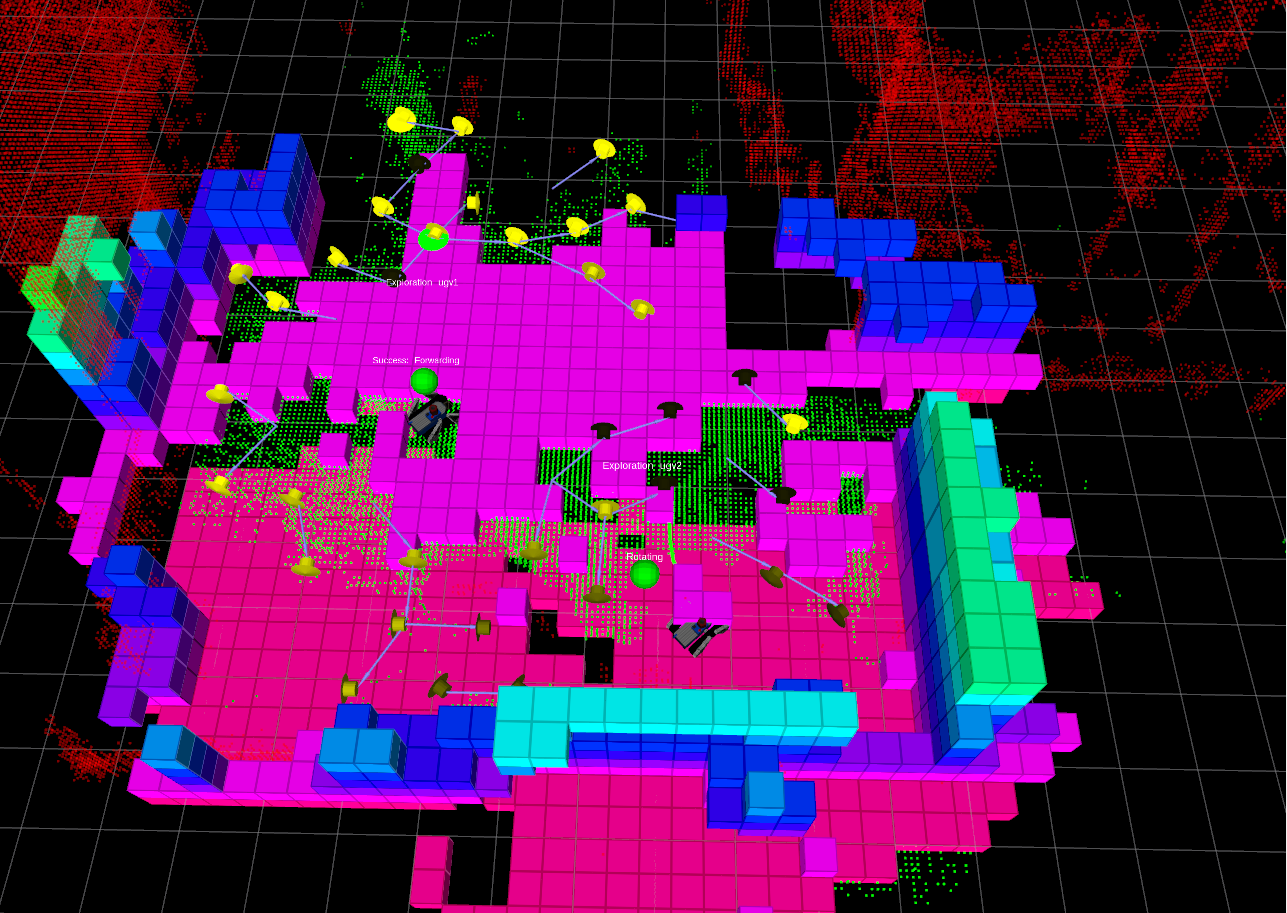}
\caption{An experiment phase at the RDM training plant. In this figure, point cloud maps and volumetric maps are overlaid. A team of two UGVs was used during the mission. The cubes represent the obstacle cells of the volumetric map ${\cal H}_j$ built by the right robot. The green points represent the segmented traversable points ${\cal M}^{trav}_j$. The red points represent the segmented obstacle points ${\cal M}^{obs}_j$. The two search trees of the robots are shown: these collect candidate view configurations, represented as small yellow arrows. In particular the lighter the yellow color of the arrows, the higher the associated utility value.}
\label{fig:exploration-deltalinq3} 
\end{center}
\end{figure}

\subsection{Evaluations with TRADR UGVs}\label{Sect:Experiments}

Our exploration framework was also tested on the TRADR UGV robots~\cite{Kruijff-2015} during some TRADR exercises in Rotterdam and Mestre. The TRADR UGVs are skid-steered and satisfy the path controllability assumption. Amongst other sensors, the robots are equipped with a $360^\circ$ spherical camera and a rotating laser scanner as shown in Fig.~\ref{Fig:Robot}. 
The full TRADR system was evaluated by the GB (Gezamenlijke Brandweer) firefighters end-users in the RDM training plant in Rotterdam (see Fig.~\ref{subfig-1:exploration-detalinq}). In particular, Fig.~\ref{subfig-2:fireman} shows the TRADR OCU (Operator Control Unit) displaying the GUI with a map and robot camera feedback.

The real multi-robot system is implemented in ROS by using a multi-master architecture. In particular, the NIMBRO network~\cite{nimbro} is used for efficiently transporting ROS topics and services over a WIFI network. Indeed, NIMBRO allows to fully leverage UDP and TCP protocols in order to control bandwidth consumption and avoid network congestion. This design choice was made by the whole TRADR consortium with the aim of developing a robust multi-robot architecture, which can flexibly offer many heterogeneous functionalities~\cite{Kruijff-2015}.

We used the same ROS C++ code in order to run both simulations and experiments. Only ROS launch scripts were adapted in order to interface the modules with the multi-master NIMBRO transport layer. 

We were able to run different exploration experiments and build a map of the plant with two UGVs. Figures \ref{fig:exploration-deltalinq1} and \ref{fig:exploration-deltalinq3} show different stages of an exploration process. During those runs the two UGVs started at the same location. In order to attain a multi-robot localization, a background process was designed to globally align the two robot maps in 3D and, in case of success, provide corrections to the mutual position estimates. This system allowed the robots to generate new maps of the facility in minutes. The pose graphs of the two registered maps were mutually consistent (up to a roto-translation) and did not present strong distributed deformations. 


%
%

\subsection{Code Implementation}\label{Sect:CodeImplementation}

All the modules are implemented in C++ and use ROS as a middleware. The code has been designed to seamlessly interface with both simulated and real robots. This allows to use the same code both in simulations and experiments.

For the implementation of our exploration agent, we used the ROS package nbvplanner as a starting point~\cite{nbvplanner}. This was specifically designed for 3D exploration with UAVs. In particular, in the accompanying paper~\cite{bircher2016receding}, the Authors present a single-robot strategy. We significantly modified the core of this package in order to manage a team of UGVs, implement our exploration agent algorithm, tightly integrate the agent module with the path planner and traversability analysis, and interface the 3D GUI in our network architecture.  

An open-source implementation is available at \href{https://github.com/luigifreda/3dmr}{github.com/luigifreda/3dmr}.

\subsection{Software Design}\label{Sect:FunctionalArchitecture}

\begin{figure*}[ht]
\centering
\includegraphics[width=0.7\linewidth]{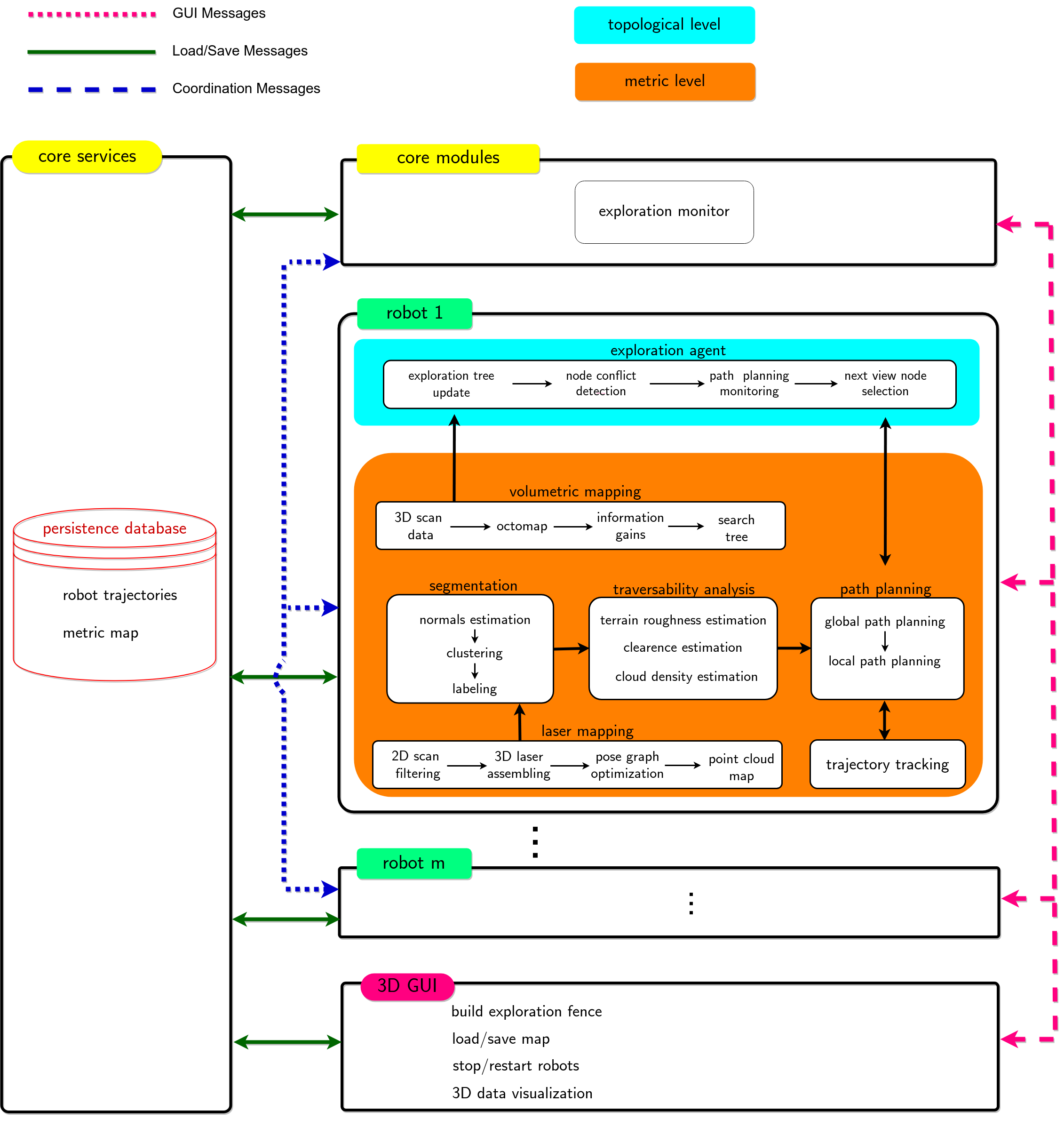}
\caption{The functional diagram of the multi-robot system. Robots share the same internal software architecture. The legend on the top left represents the different kind of exchanged messages.}
\label{Fig:Architecture}
\end{figure*}

A functional diagram of the presented multi-robot system is reported in Fig.~\ref{Fig:Architecture}. The main blocks are listed below.

The robots, each one with its own ID~$\in \{1,...,m\}$, have the same internal architecture and host the on-board functionalities that concern decision and processing aspects both at the topological level and at the metric level. 

The core services, hosted in the main central computer, represent the multi-robot system persistence. These allow specific modules to load and save maps and robot trajectories (along with other TRADR data structures).

The core modules, also hosted in the central computer, include the exploration monitor. This continuously checks the current status of the exploration activities and records relevant data.

The multi-robot 3D GUI hosted on one OCU is based on \texttt{rviz} and allows the user \emph{(i)} to build an \emph{exploration fence}, which consists of a set of points defining a polygonal bounding region and limiting the explorable area \emph{(ii)} to visualize relevant point cloud data, maps, and robot models \emph{(iii)} to stop/restart robots when needed \emph{(iv)} to trigger the loading/saving of maps and robot trajectories.

Each robot hosts its instance of the exploration agent and of the path-planner. Hence, the implemented exploration algorithm is fully distributed. 

As illustrated in Fig.~\ref{Fig:Architecture}, the various modules in the architecture exchange different kinds of messages. These can be mainly grouped into the following types.
\begin{itemize}
\itemsep0em 
\item Coordination messages: these are mainly exchanged amongst robots  in order to achieve coordination and cooperation. For convenience, the exploration monitor records an history of these messages. 
\item GUI messages: these are exchanged with the 3D GUI and include both control messages and visualization data.
\item Load/save messages: these are exchanged with the core services and contain both loaded and saved data.
\end{itemize}

\section{Conclusions}\label{Sect:Conclusions}

We have presented a 3D multi-robot exploration framework for a team of UGVs navigating in unknown harsh terrains with possible narrow passages. We cast the two-level coordination strategy presented in \cite{fre2018patrolling} in the context of exploration. An analysis of the proposed method through both simulations and real-world experiments was conducted~\cite{Novo2021}. 

We extended the receding-horizon next-best-view approach~\cite{bircher2016receding} to the case of UGVs by suitably adapting the expansion of sampling-based trees directly on traversable regions. In this process, an iterative windowed search strategy was used to optimize the computation time. In our tests, the proposed frontier trees and backtracking strategy proved to be crucial to effectively complete exploration experiments without leaving unvisited regions. 
The approach of separating the resolution of topological and metric conflicts showed to be successful also in this context~\cite{fre2018patrolling}. The resulting strategy is distributed and succeeds to minimize and explicitly manage the occurrence of conflicts and interferences in the robot~team. 

A prioritization scheme was integrated into the framework to steer the robot exploration toward areas of interest on-demand. This design was suggested by TRADR end-users to improve their control of the exploration progress and better integrate human and robot decision-making during a real mission. The presented framework can be used to perform coverage tasks in the case a map of the environment is a priori provided as~input.

The conducted experiments proved that our exploration method has a competitive performance and allows appropriate management of a team of robots. Decentralization allows task reallocation and recovery when robot failures occur. Notably, the proposed approach showed to be agnostic and not specific to a given sensing configuration or robotic platform. 

We publish the source code of the presented framework
with the aim of providing a useful tool for researchers in the
Robotics Community. In the future, we plan to improve the system communication protocol and to validate the framework on larger teams of robots. Moreover, we aim to conduct a comprehensive comparison and performance evaluation of the proposed exploration method against other state-of-the-art 3D exploration approaches under the same conditions.

%


\bibliographystyle{IEEEtran}
\bibliography{biblio/biblio,biblio/bibliography}

\end{document}